\documentclass[journal]{IEEEtran}

\usepackage{bm}
\usepackage{cite}
\usepackage{algorithm}
\usepackage{algorithmicx}
\usepackage{amsmath}
\usepackage{algpseudocode}
\usepackage{subfigure}
\usepackage{graphicx}
\usepackage{multirow}
\usepackage{color, xcolor}

\usepackage{tcolorbox}
\usepackage{xspace}

\renewcommand{\baselinestretch}{0.94}

\newcommand{\BULLET}{\vspace{+.00in} \noindent $\bullet$ \hspace{+.00in}}

\newcommand{\ie}{\textit{i.e.,}\xspace}

\begin{document}

\title{An Efficient Batch Constrained Bayesian Optimization Approach for Analog Circuit Synthesis via Multi-objective Acquisition Ensemble}



\author{Shuhan Zhang, Fan Yang, Changhao Yan, Dian Zhou, Xuan Zeng \vspace{-0.3in}
\thanks{
    The preliminary version has been presented at International Conference on Machine Learning (ICML) in 2018.
    This research is supported partly by National Key R\&D Program of China 2020YFA0711900, 2020YFA0711901, National Natural Science Foundation of China (NSFC) research projects 61822402, 61774045, 61974032, 61929102 and 62011530132.}
\thanks{*Corresponding authors: Fan Yang; Xuan Zeng. Email: \{yangfan, xzeng\}@fudan.edu.cn}
\thanks{S.~Zhang, F.~Yang, C.~Yan and X.~Zeng are with State Key Lab of ASIC \& System, Microelectronics Department, Fudan University, China.}
\thanks{D.~Zhou is with the Department of Electrical Engineering, The University of Texas at Dallas, USA.}
}

\maketitle

\begin{abstract}
    Bayesian optimization is a promising methodology for analog circuit synthesis. However, the sequential nature of the Bayesian optimization framework significantly limits its ability to fully utilize real-world computational resources. In this paper, we propose an efficient parallelizable Bayesian optimization algorithm via Multi-objective ACquisition function Ensemble (MACE) to further accelerate the optimization procedure. By sampling query points from the Pareto front of the probability of improvement (PI), expected improvement (EI) and lower confidence bound (LCB), we combine the benefits of state-of-the-art acquisition functions to achieve a delicate tradeoff between exploration and exploitation for the unconstrained optimization problem. Based on this batch design, we further adjust the algorithm for the constrained optimization problem. By dividing the optimization procedure into two stages and first focusing on finding an initial feasible point, we manage to gain more information about the valid region and can better avoid sampling around the infeasible area. After achieving the first feasible point, we favor the feasible region by adopting a specially designed penalization term to the acquisition function ensemble. The experimental results quantitatively demonstrate that our proposed algorithm can reduce the overall simulation time by up to $74\times$ compared to differential evolution (DE) for the unconstrained optimization problem when the batch size is 15. For the constrained optimization problem, our proposed algorithm can speed up the optimization process by up to $15\times$ compared to the weighted expected improvement based Bayesian optimization (WEIBO) approach, when the batch size is 15.


\end{abstract}


\IEEEpeerreviewmaketitle

\section{Introduction} \label{sec:introduction}



With the scaling integrated circuit (IC) technology, circuit devices are becoming more complex and the parasite effect can no longer be ignored, which in turn complicates circuit design. Although digital circuit design has been automated for a long time, analog circuits are still designed manually. Due to the growing demands for high-performance, low-power and short-to-market circuits, designing analog circuits manually has become increasingly intractable. Therefore, automated analog circuit design tools are in urgent need. 

Analog circuit design consists of two steps: topology selection and device sizing. In this paper, we focus on the device sizing problem, which can be formulated into both unconstrained and constrained optimization problem (\S\ref{sec:problem_formulation}). Since the computational cost of circuit simulations can be prohibitively expensive and sometimes intractable, the evaluation budget to search for the optimum circuit design should be kept at a minimum level. What's worse, the derivatives and the convexity property of the circuit sizing problem are always inaccessible. Thus, the corresponding optimization algorithm should be able to handle the problems that are costly, noisy and multi-modal. 

Decades of scientific efforts have been devoted to developing efficient optimization algorithms, which generally fall into two categories: \textit{model-based} and \textit{simulation-based} methods. The model-based approaches try to speed up the optimization process by constructing a cheap-to-evaluate substitute for the circuit simulation. Designers prescribe their prior knowledge about the analog circuit and manually derive the analytical expression to approximate the circuit performance. The automated regression algorithms will also be taken into consideration when the analytical expression can not be derived or not available \cite{rutenbar2007hierarchical}. One influential model-based approach is the geometric programming algorithm \cite{boyd2005geometric, boyd2007tutorial, hannah2012ensemble}, which models the circuit performance with posynomial approximation. Other modeling strategies also exist, including artificial neural network (ANN) \cite{jafari2010design, li2019artificial, rosa2019using}, support vector machine (SVM) \cite{li2020nonlinear}, and Gaussian process regression (GPR) \cite{he2020efficient, lyu2017efficient, zhang2019bayesian, zhang2019efficient, zhang2020bayesian, pan2019analog}. The disadvantage that prevents model-based methods from being widely used is that an accurate performance model is always hard to derive manually or requires a large set of simulation data to approximate. Besides, the generated model is not guaranteed to be accurate all over the design space. Considering that the constructed model may deviate from the real circuit performance, the optimization results can also divert from the real optimum.


The simulation-based approaches, on the other hand, view the circuit performance as a black-box function and optimize it on the fly. By leveraging the previously observed dataset, the simulation-based approaches guide the search by proposing the potential locations for evaluation. Since the computational cost of circuit simulations can be prohibitively expensive, the required number of evaluations should be kept at a minimum level to accelerate the optimization process. Embodiments of simulation-based approaches include the simulated annealing (SA) \cite{gielen1990analog}, the evolutionary algorithm \cite{alpaydin2003evolutionary,liu2009analog, liu2011efficient} \cite{phelps2000anaconda:}, the multiple start points (MSP) algorithm \cite{yang2018smart-msp:}\cite{peng2016efficient}, and the particle swarm optimization (PSO) algorithm \cite{fakhfakh2010analog,wu2009a,coello2003use}. All of these proposed algorithms try to mimic the physical or biological process to fully explore the state space and avoid stuck in the local optimum. The biggest limitation that hinders the simulation-based approaches from being widely used is their relatively low convergence rate.


Inspired by the above, the Bayesian optimization framework has been proposed to fully accelerate the optimization process by \textit{combining} the model-based and simulation-based approaches \cite{shahriari2015taking,liu2014gaspad:,lyu2017efficient,lyu2018batch,lyu2018multi,zhang2019bayesian,zhang2019efficient,zhang2020bayesian,he2020efficient,hu2018parallelizable,wangmulti,liu2016multi}. It is quite suitable for problems that don't have a closed-form expression for the objective function and can only be observed through sampled values \cite{shahriari2015taking}. The Bayesian optimization algorithm is especially efficient in situations when the sampled values are noisy, evaluations are incredibly expensive, or the convexity properties are unknown. Generally, there are two key elements in the Bayesian optimization framework: the probabilistic surrogate model and the acquisition function (\S\ref{sec:background}). The surrogate model incorporates our prior belief and provides a posterior distribution with the observed dataset. The prescribed prior belief is the modeling space of the possible latent function. The posterior distribution means the surrogate model provides not only the predictive means but also the corresponding uncertainty estimations. In other words, the surrogate model works as a cheap-to-evaluate substitute for the expensive latent function. The acquisition function instead describes the data generation mechanism. By leveraging the provided posterior distribution, it works as a utility-based selection criterion that helps to direct the sampling process. Instead of solely searching for the global optimum over the predictive mean, the acquisition function takes both exploration and exploitation into consideration. This means the acquisition function favors the potential region with high uncertainty estimations (exploration) and the area that is predicted to be optimal with high probability (exploitation). In this way, the acquisition function can better explore the state space and help to select better candidate points. As opposed to the model-based approaches that explore the design space offline, the Bayesian optimization algorithm updates the observed dataset incrementally and refines the surrogate model to provide a more informative posterior distribution at each iteration. It also gives us a theoretically-guaranteed global optimum after a certain number of observations. In summary, the Bayesian optimization algorithm opens a more fast and efficient lane for global optimization. Thanks to the Bayesian optimization framework, we can bypass the traffic jams of the model-based methods that greatly depend on the accuracy of the constructed model and the simulation-based approaches that have relatively low convergence rate.

Although the Bayesian optimization algorithm has been widely used to search the state space \cite{he2020efficient,zhang2019bayesian,zhang2019efficient,lyu2017efficient}\cite{liu2014gaspad:}\cite{lyu2018multi}\cite{liu2016multi}, the sequential decision-making nature of the acquisition function prevents it from being parallelized. Without parallelism, the computational resources are not fully utilized and the optimization efficiency is greatly limited, especially in situations when the multi-core workstations are available and the circuit simulations are computationally intensive. To further tap the potential of the Bayesian optimization framework, great efforts have been made to make parallelism possible, which means the algorithm can propose several data points at each iteration. The simulation process can be distributed to different workers on the workstation. However, there are two challenges in parallelizing the Bayesian optimization algorithm. The first one is to avoid sampling redundantly around the same region, since maximizing the existing state-of-the-art acquisition functions naturally selects around the same region. The second one is to maximize the information gain of each query point in the batch. To solve these problems, most of the state-of-the-art batch Bayesian optimization algorithms design a penalization scheme that can penalize around the previously selected query points and select the candidate points in a batch one by one, like local penalization (LP) strategy \cite{gonzalez2016batch}, batched Bayesian optimization algorithm based on the lower confidence bound (BLCB) \cite{desautels2014parallelizing}, parallelizable Gaussian process optimization with upper confidence bound and pure exploration (GPUCB-PE) \cite{Contal2013Parallel}, and parallelizable Bayesian optimization algorithm with high coverage consideration (pHCBO) \cite{hu2018parallelizable}.
Another limitation of the existing batch Bayesian optimization approaches is that they solely rely on a single acquisition function. 
Although there exist batch Bayesian optimization algorithms that can use arbitrary acquisition functions \cite{gonzalez2016batch}\cite{azimi2010batch} to facilitate the query points selection in a batch, most of the state-of-the-art batch Bayesian optimization algorithms including BLCB \cite{desautels2014parallelizing} and GPUCB-PE \cite{Contal2013Parallel} depend solely on one acquisition function, which can greatly limit their performance.


Although a large body of scientific literature has been published on developing a well-designed batch Bayesian optimization algorithm \cite{lyu2018batch}\cite{desautels2014parallelizing,Contal2013Parallel,azimi2010batch}, almost all these works solely focus on the unconstrained optimization problem -- the constrained optimization problem is rarely considered. 
In this paper, we propose an efficient batch Bayesian optimization algorithm that can handle both unconstrained and constrained optimization problems. By randomly sampling data points from the Pareto front of the state-of-the-art acquisition functions, our proposed algorithm naturally maintain diversity in a batch while maximizing the information gain per observation for the unconstrained optimization problem. For the constrained optimization problem, we divide the optimization process into two stages to better explore the design space: (1) seeking the first feasible point and (2) searching for the global optimum that satisfies the constraints. In this way, we can focus on finding the first valid point before taking both constraints and objective into consideration. A preliminary version of this paper was published in \cite{lyu2018batch}.
We test our algorithm on four real-world analog circuits to quantitatively demonstrate the efficiency and effectiveness of our proposed algorithm. For the unconstrained optimization problem, our proposed algorithm can reduce the simulation time by up to $74\times$ compared to DE \cite{liu2009analog} when the batch size is 15. For the constrained optimization problem, our proposed algorithm can speed up the optimization process by up to $15\times$ compared to WEIBO \cite{lyu2017efficient} when the batch size is 15.

The remainder of this paper is organized as follows. Section \S\ref{sec:problem_formulation} gives the problem formulation of the analog circuit device sizing problem. Section \S\ref{sec:background} introduces the background knowledge of our proposed algorithm. Section \S\ref{sec:proposal} presents the fundamental challenges of parallelizing Bayesian optimization framework and the MACE algorithm for the unconstrained optimization problem. Section \S\ref{sec:proposal} explicitly presents our improved batch constrained Bayesian optimization algorithm and the spirit behind it. In section \S\ref{sec:experiment}, we report the experimental results and analytically compare the performances of our proposed algorithm with the state-of-the-art optimization algorithms. We conclude the paper in section \S\ref{sec:conclusion}.


\section{Problem Formulation}
\label{sec:problem_formulation}

In this section, we formulate the analog circuit device sizing problem into the unconstrained optimization problem (\S\ref{sec:unconstrained}) and the constrained optimization problem (\S\ref{sec:constrained}).

\subsection{Unconstrained Optimization Problem}
\label{sec:unconstrained}

For a given circuit topology, the analog circuit optimization problem can be formulated as an unconstrained optimization problem by combining several circuit performances with weighting parameters:
\begin{equation}
	\label{eq:unconstrained}
	\text{minimize} \quad \text{FOM} = \sum^{M}_{i=1} \alpha_i f_i(\bm{x}),
\end{equation}
where $\bm{x} \in R^d$ represents the input vector constructed by $d$ design variables, $f_i(\cdot)$ stands for the $i$-th performance matric of $M$ circuit performances, $\alpha_i$ is the $i$-th weighting parameter, and FOM denotes our interested Figure of Merit. For simplicity of denotation, we only consider the minimization problem in this paper.



\subsection{Constrained Optimization Problem} \label{sec:constrained}

The analog circuit device sizing problem can also be formulated as a constrained optimization problem:
\begin{equation}
	\label{eq:constrained}
	\begin{aligned}
		& \text{minimize} & & \text{FOM} \\
		& \text{s.t.} & & c_i(\bm{x}) < 0 \\
		& & & \forall i \in {1 \dots N_c},
	\end{aligned}
\end{equation}
where $N_c$ denotes the number of constraints, and $c_i(\cdot)$ is the $i$-th constraint function. The target of the circuit design is to search for a circuit design that minimizes the FOM while satisfying constraints.




\section{Review of Bayesian Optimization} \label{sec:background}


In this section, we introduce the background knowledge of our proposed algorithm and give a brief overview of the Bayesian optimization framework based on the Gaussian process regression model in \S\ref{sec:bayesian_optimization}. We also outline several state-of-the-art acquisition functions and highlight their selection principles in \S\ref{sec:acquisition_function}.




\subsection{Bayesian Optimization} 
\label{sec:bayesian_optimization}

\begin{algorithm}
	\caption{Bayesian Optimization Framework}
	\label{algo:bayesian_optimization}
	\begin{algorithmic}[1]
		\Require The size of the initial dataset $N_{init}$, and the maximum number of iteration $N_{iter}$
		\State Randomly sample a initial dataset $D_{0}=\{X, \bm{y}\}$
		\For{$t=0 \to N_{iter}$}
		\State Construct a Gaussian process regression model with $D_t$
		\State $\bm{x}_t \gets \text{argmax}_{\bm{x}} \alpha(\bm{x}; D_t)$
		\State $y_t = f(\bm{x}_t)$
		\State $D_{t+1} \gets \{D_t, \{\bm{x}_t, y_{t}\}\}$
		\EndFor
	    \State \Return Best $y$ recorded after optimization
	\end{algorithmic}
\end{algorithm}

Bayesian optimization framework has demonstrated significant potential in approximating the global optimum with a relatively small number of evaluations \cite{lyu2017efficient,lyu2018batch,lyu2018multi,zhang2019bayesian,zhang2019efficient,zhang2020bayesian,hu2018parallelizable,desautels2014parallelizing}. It gains efficiency by leveraging both the surrogate model and the acquisition function \cite{shahriari2015taking}. The surrogate model works as a simplified representation of the costly simulation process by taking the whole history of optimization into considerations. The informative posterior distribution provided by the surrogate model includes the predictive mean and well-calibrated uncertainty estimation. The acquisition function prioritizes data points in the candidate pool and guides the search by proposing a sequence of promising data points. A comprehensive review of the Bayesian optimization framework can be found in \cite{shahriari2015taking}, and the corresponding framework is presented in Algorithm \ref{algo:bayesian_optimization}.


The Gaussian process regression model is one of the most commonly used surrogate models in the Bayesian optimization framework \cite{rasmussen2003gaussian}. Given a $d$-dimensional input design variable $\bm{x}$, we assume the unknown objective function as $y=f(\bm{x})+\epsilon$, where $\epsilon$ denotes the observation noise $N(0, \sigma^2_n)$. Let us assume the accumulated observations as $D=\{X, \bm{y}\}$, where $X$ represents a set of design variables $X=\{\bm{x}_1, \bm{x}_2, \cdots, \bm{x}_N\}$, and $\bm{y}$ denotes the corresponding $N$ observations $\bm{y}=\{y_1, y_2, \cdots, y_N\}$. By capturing our prior belief about the performances of the unknown objective function with predefined mean function $m(\bm{x})$ and kernel function $k(\bm{x}_i, \bm{x}_j)$, the Gaussian process regression model can provide posterior distribution for an arbitrary location $\bm{x}^\ast$ as follow \cite{rasmussen2003gaussian}:
\begin{equation}
	\begin{cases}
		\mu(\bm{x}^\ast) = k(\bm{x}^\ast, X)[K+\sigma^2_nI]^{-1}\bm{y} \\
		\sigma^2(\bm{x}^\ast) = k(\bm{x}^\ast, \bm{x}^\ast) - k(\bm{x}^\ast, X)[K+\sigma^2_nI]^{-1}k(X, \bm{x}^\ast),
	\end{cases}
\end{equation}
where $\mu(\bm{x}^\ast)$ is the predictive mean, $\sigma(\bm{x}^\ast)$ denotes the uncertainty estimation, $k(\bm{x}^\ast, X) = k^{T}(X, \bm{x}^\ast)$, and $K=k(X, X)$ is the corresponding covariance matrix. In this paper, we set $m(\bm{x})=0$ and the kernel function as the squared exponential (SE) covariance function:
\begin{equation}
	\label{eq:squared_exponential}
	k_{SE}(\bm{x}_i, \bm{x}_j) = \sigma_f^2 \text{exp}(-\frac{1}{2} (\bm{x}_i - \bm{x}_j)^T \Lambda^{-1} (\bm{x}_i - \bm{x}_j)).
\end{equation}
In equation (\ref{eq:squared_exponential}), $\sigma_f$ denotes the variance, $\Lambda = \text{diag}(l_1, \cdots, l_d)$ is a $d \times d$ diagonal matrix, and $l_i$ represents the length scale of the $i$-th dimension. The hyperparameter vector $\bm{\theta} = (\sigma_n, \sigma_f, l_1, \cdots, l_d)$ can be determined during the model training process. For more detailed discussion about Gaussian processes, we refer readers to \cite{rasmussen2003gaussian}.

\subsection{Acquisition Function} \label{sec:acquisition_function}

The acquisition function in the Bayesian optimization algorithm works as a cheap-to-evaluate utility function to guide the sampling decisions. Instead of exploring the design space only with the predictive mean, the acquisition function leverages the uncertainty estimation to explore the unknown area, until they are confidently ruled out as suboptimal. In this way, the acquisition function favors not only the current promising area with high confidence but also the unknown region with large uncertainty estimation. In other words, the acquisition function trades off between the exploration and exploitation based on the posterior beliefs provided by the surrogate model. There are three most widely used acquisition functions.


\textbf{The Probability of Improvement (PI).} Given the current minimum objective function value $\tau$ in the dataset, the probability of improvement function tries to measure the probability an arbitrary $\bm{x}$ exceeds the current best. The corresponding formulation is as follow \cite{kushner1964new}:
\begin{equation}
	\label{eq:PI}
	\text{PI}(\bm{x}) = \Phi(\lambda),
\end{equation}
where $\Phi(\cdot)$ is the cumulative distribution function (CDF) of standard normal distribution. Following the suggestion of \cite{brochu2010tutorial}, we introduce a small positive jitter $\xi$ to encourage exploration and set $\lambda = (\tau - \xi - \mu(\bm{x})) / \sigma(\bm{x})$. In this paper, we fix $\xi$ as 0.001.

\textbf{The Expected Improvement (EI).} Compared with PI that only measures the probability of improvement and treats the improvement equally, the expected improvement function tries to measure the amount of improvement upon the current best $\tau$. By maximizing the expected improvement function, we can expect that the observation $\bm{x}$ will not only exceed the current best but also exceed the current best value at the highest magnitude. The corresponding formulation can be expressed as \cite{mockus1978application}:
\begin{equation}
	\label{eq:EI}
	\text{EI}(\bm{x}) = \sigma(\bm{x}) (\lambda \Phi(\lambda) + \phi(\lambda)),
\end{equation}
where $\phi(\cdot)$ is the probability density function (PDF) of standard normal distribution.

\textbf{The Lower Confidence Bound (LCB).} Compared with the improvement-based strategies like PI and EI, the lower confidence bound function tries to guide the search from an optimistic perspective. With the carefully designed coefficient $\beta$, the cumulative regret is theoretically bounded \cite{srinivas2009gaussian} \cite{srinivas2012information}. Thus, the convergence of the Bayesian optimization algorithm is guaranteed. The corresponding formulation is:
\begin{equation}
	\label{eq:LCB}
	\text{LCB}(\bm{x}) = \mu(\bm{x}) - \beta \sigma(\bm{x}).
\end{equation}
In this paper, we follow the suggestion of \cite{brochu2010tutorial} and set $\beta = \sqrt{2 \nu log(t^{d/2+2} \pi^2 / 3 \delta)}$, where $t$ denotes the number of iterations, $\nu$ and $\delta$ are two user-defined parameters. In this paper, we fix $\nu=0.5$ and $\delta=0.05$. By minimizing the LCB function, the global optimum can be achieved within a limited number of observations.

Apart from the above mentioned acquisition functions, there are also some other types of acquisition functions, including entropy search (ES) \cite{hennig2012entropy}, Thompson sampling (TS) \cite{thompson1933likelihood}, predictive entropy search (PES) \cite{hernandez2014predictive}, max-value entropy search (MES) \cite{wang2017max} and knowledge gradient (KG) \cite{scott2011correlated, frazier2009knowledge}. It is also possible to explore the state space with a portfolio of acquisition functions \cite{hoffman2011portfolio, shahriari2014entropy}.

\section{Batch Bayesian Optimization} \label{sec:proposal}





In this section, we first identify the fundamental challenges in parallelizing Bayesian optimization algorithm (\S\ref{sec:batch_BO}). We then explicitly describe our batch algorithm design and the spirit behind it (\S\ref{sec:MACE}). Finally, we briefly review the multi-objective algorithm that we use to facilitate the optimization procedure (\S\ref{sec:MO}).



\subsection{Challenges for Batch Bayesian Optimization}
\label{sec:batch_BO}

Since the traditional state-of-the-art acquisition functions generally work as a utility function and help to explore the design space by prioritizing the candidate data points, they naturally work in sequential mode and tend to select the same location repetitively. Due to the information gap between decisions and observations, classical works on batch Bayesian optimization generally address this problem by penalizing around the previous selections and sampling query points in a batch one by one. Compared to sequential Bayesian optimization, there are two fundamental challenges for designing an efficient parallelizable Bayesian optimization algorithm: 

\BULLET\textbf{C1:} How to maximize the information gain of each data point?

\BULLET\textbf{C2:} How to maintain high diversity within each batch?

To tackle \textbf{C1}, most of the existing batch Bayesian optimization algorithms simply guide the search with traditional acquisition functions like EI, PI and LCB. However, considering that no single acquisition function can always outperform others \cite{hoffman2011portfolio}, searching the design space by relying on solely one acquisition function can greatly limit the efficiency of the optimization procedure.


To address \textbf{C2}, most of the existing batch Bayesian optimization algorithms introduce a carefully designed penalization scheme to reduce sampling around the same region redundantly. For example, LP \cite{gonzalez2016batch} proposes to penalize around the previous decisions in the batch with a manually designed local penalization strategy by introducing the Lipschitz constant as local repulsion. BLCB \cite{desautels2014parallelizing} instead heuristically encourages diversity in a batch with fake observations. Since the uncertainty estimations for arbitrary locations do not depend on the objective values, it penalizes around the previous decisions in a batch by taking advantages of the modeling property of the Gaussian process regression model. GPUCB-PE \cite{Contal2013Parallel} is an exploratory batch design that combines the benefits of the upper confidence bound (UCB) policy and the pure exploration strategy to improve the selection efficiency from a theoretical perspective. pBO \cite{hu2018parallelizable} proposes to select the batch with different weighting parameters to balance between the exploration and exploitation and encourage diversity in a batch. Based on pBO, pHCBO \cite{hu2018parallelizable} further introduces a specially designed penalization scheme to prevent cluster sampling by the same weighting parameter.

In other words, the solution to \textbf{C2} adopted by major batch designs is to propose locations for evaluation by mimicking the sequential process. Specifically, to maximize the payoff of each observation, the state-of-the-art batch policies reduce redundantly sampling over the same region and marginalize around the previous elements in a batch by introducing a carefully designed local repulsive term. Nonetheless, this solution is not good enough. The problem is that \textit{a manually designed penalization scheme always introduces human biased presumptions into the batch selection procedure, thus, tends to be overconfident about the decision-making process.} For LP \cite{gonzalez2016batch}, the Lipschitz constant itself requires lots of effort to approximate and the corresponding approximation can deviate from its real value, thus, hinders the efficiency of the selection procedure. For BLCB \cite{desautels2014parallelizing}, the batch policy generally assumes the objective value of each previous selection equals to the predictive mean, which deteriorates the performance of BLCB for problems that have a quick changing response surface. For GPUCB-PE \cite{Contal2013Parallel}, the specially designed pure exploration procedure focuses on reducing the systematic uncertainty and is not efficient enough for searching the global optimum. The penalization scheme proposed by pHCBO \cite{hu2018parallelizable} is not efficient enough, since the refined surrogate model itself will provide a reduced uncertainty estimation for the new batch design and will naturally penalize around the previous batch.

We next describe how to truly overcome \textbf{C1} and \textbf{C2} by detailing our batch algorithm design.

\subsection{Batch Bayesian Optimization via Acquisition Function Ensemble} \label{sec:MACE}


\begin{figure}
    \centering
    \includegraphics[width=0.45\textwidth]{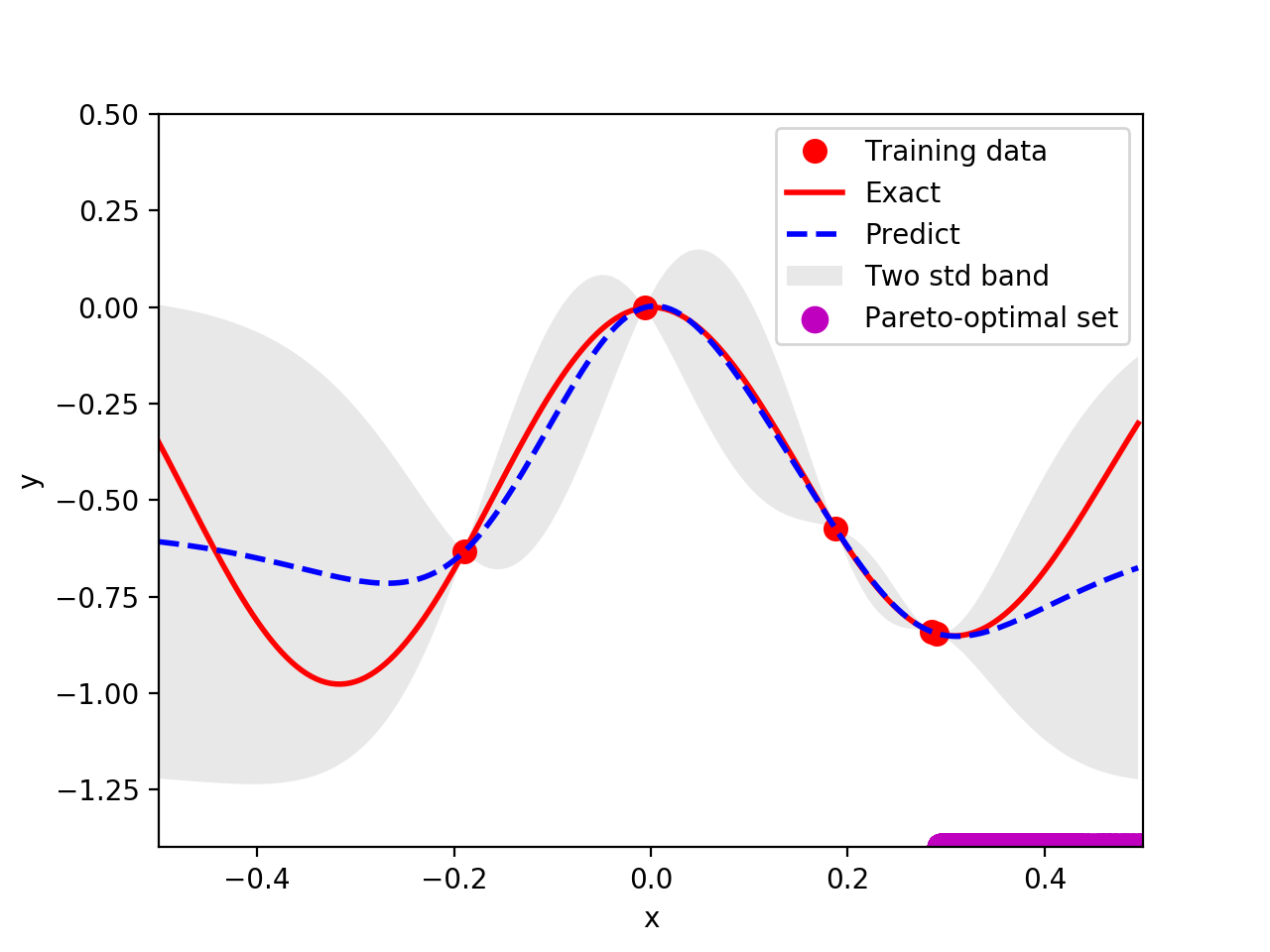}
    \caption{An illustration of the posterior distribution provided by the Gaussian process regression model and the corresponding Pareto-optimal set of the multi-objective acquisition function ensemble.}
    \label{fig:acquisition_function}
\end{figure}

The behavior of space exploration for a Bayesian optimization algorithm is largely determined by its acquisition function. Different acquisition functions are designed under the guidance of different selection principles, thus, tend to propose different candidate points for evaluation. Each acquisition function has its strengths and weaknesses.
For example, PI selects data points that maximize the probability of improvement and is naturally biased towards exploitation. For problems with multiple local optima, PI tends to be too conservative and is not efficient enough for exploring the design space. EI instead takes the magnitude of improvement into consideration and guides the search in a greedy manner. By pickling data points that are compatible with the current best, EI tries to obtain improvement in each step. However, due to its greedy-choice property, EI function is not guaranteed to guide the search towards the right direction and convergence to the global optimum for a given number of iterations. LCB guides the search by minimizing the cumulative regret bound and theoretically guarantees the convergence after a relative number of iterations. But the exact value of the hyperparameter $\beta$ is hard to define, since the requirement for exploration and exploitation could differ for different problems. For a given evaluation budget, LCB is also not guaranteed to achieve the best optimization results. In other words, no acquisition function works best in every scenario and \textit{no free lunch} in optimization \cite{hoffman2011portfolio, wolpert1997no}. Therefore, searching the design space by solely relying on a single acquisition function can greatly limit the efficiency and effectiveness of the optimization procedure.



To fill this gap, we propose a parallelizable Bayesian optimization algorithm based on the Multi-objective ACquisition function Ensemble (MACE). By sampling data points simultaneously from the Pareto front of PI, EI and LCB, we combine the benefits of the state-of-the-art acquisition functions and capture the best tradeoff between exploration and exploitation (\textbf{C1}). In this way, we can maximize the information gain of each selected data point without introducing human biased knowledge, while naturally encouraging diversity within the batch even without penalization scheme (\textbf{C2}). Other acquisition functions like KG and ES can also be incorporated into MACE framework to facilitate the space exploration procedure. The corresponding formulation is as follow:
\begin{equation}
    \label{eq:MACE}
    \text{minimize} \quad \text{LCB}(\bm{x}), -\text{PI}(\bm{x}), -\text{EI}(\bm{x}).
\end{equation}
In Figure \ref{fig:acquisition_function}, we give an illustration of the proposed Pareto-optimal set of the multi-objective acquisition function ensemble. The overall framework of the proposed parallelizable Bayesian optimization algorithm via multi-objective acquisition function ensemble is presented in Algorithm \ref{algo:MACE}. Compared with most of batch Bayesian optimization algorithms that select data points one by one, our parallelizable Bayesian optimization framework naturally maintains diversity within each batch and simultaneously selects candidate points as a whole. 

\begin{algorithm}
    \caption{MACE algorithm}
    \label{algo:MACE}
    \begin{algorithmic}[1]
        \Require The size of the initial dataset $N_{init}$, the maximum number of iteration $N_{iter}$, and the batch size $B$.
        \State Randomly sample an initial dataset $D_0=\{X, \bm{y}\}$
        \For{$t = 0 \to N_{iter}$}
        \State Construct a Gaussian process regression model with training dataset $D_t$
        \State Generate a Pareto-optimal dataset $P_t$ with equation (\ref{eq:MACE})
        \State Randomly sample $B$ data points $X_t=\{\bm{x}_{t,1}, \bm{x}_{t,2}, \cdots, \bm{x}_{t, B}\}$ from $P_t$
        \State Evaluate the sampled data points $\bm{y}_t = \{f(\bm{x}_{t,1}), f(\bm{x}_{t,2}), \cdots, f(\bm{x}_{t,B}))\}$
        \State $D_t \gets \{D_{t-1}, \{X_t, \bm{y}_t\}\} $
        \EndFor
        \State \Return Best $y$ recorded after optimization
    \end{algorithmic}
\end{algorithm}

\subsection{Multi-objective Optimization} \label{sec:MO}

To obtain the \textit{Pareto-optimal set} of the acquisition function ensemble, we introduce the multi-objective optimization algorithm to facilitate the optimization procedure. The multi-objective optimization algorithms aim to optimize several objective functions at the same time, which can be expressed as:
\begin{equation}
	\label{eq:multi-objective}
	\text{minimize} \quad f_1(\bm{x}), f_2(\bm{x}), \cdots, f_m(\bm{x}).
\end{equation}
Unlike the single-objective optimization, there is usually no single data point that has the best performances for all objective functions and the objective functions can contradict with each other. Instead of getting a single global optimum, we can only generate a \textit{Pareto-optimal set} that no data point dominates any other. For arbitrary data points $\bm{a}$ and $\bm{b}$, we call $\bm{a}$ dominates $\bm{b}$ if:
\begin{equation}
	\label{eq:dominant}
    \begin{aligned}
	& \forall i \in \{1,\cdots, m\} f_i(\bm{a}) \geq f_i(\bm{b}) & \land \\
	& \exists i \in \{1, \cdots, m\} f_i(\bm{a}) > f_i(\bm{b}). &  
	\end{aligned}
\end{equation}
We refer the entire non-dominated design space as the \textit{Pareto-optimal front}.

A large body of literature has been published on multi-objective optimization algorithms, including non-dominated sorting genetic algorithm II (NSGA-II) \cite{deb2002fast}, efficient global optimization algorithm with Gaussian processes model (ParEGO) \cite{knowles2006parego}, the multi-objective evolutionary algorithm based on decomposition (MOEA/D) \cite{zhang2007moea}, the improved strength Pareto evolutionary algorithm (SPEA2) \cite{zitzler2001spea2}, the specially designed multi-objective particle swarm optimization (OMOPSO) algorithm \cite{sierra2005improving}, the speed-constrained multi-objective particle swarm optimization algorithm (SMPSO) \cite{nebro2009smpso}. In this paper, we use differential evolution for multi-objective optimization (DEMO) algorithm \cite{robivc2005differential} to obtain the Pareto-optimal set of the proposed acquisition function and facilitate the optimization procedure. In this paper, we fix the population size as 100 and the number of evaluations as 2000 during the acquisition function optimization procedure.

\section{Constrained Batch Bayesian Optimization} \label{sec:constrained_proposal}

Due to the limitations in real-world circuit design, some circuit performances should be kept below a certain level. Therefore, we propose a refined MACE algorithm to handle this constrained optimization problem. To fully maximize the information gain for both objective and constraints, we divide the optimization procedure into two stages: (1) seeking the first feasible point (\S\ref{sec:stage_one}), (2) searching for the global optimum that satisfies constraints (\S\ref{sec:stage_two}). \S\ref{sec:summary} summarizes our proposed batch constrained Bayesian optimization algorithm.

\vspace{-0.1in}
\subsection{Seek the First Feasible Point} \label{sec:stage_one}


For the constrained problem, we decide to first focus on finding the first feasible point when there is no feasible point in the dataset. In this way, we can not only obtain the first feasible point more quickly but also provide more information about the feasible region for further optimization. One of the most widely used acquisition functions to favor the feasible region is the probability of feasibility (PF):
\begin{equation}
    \label{eq:PF}
    \text{PF}(\bm{x}) = \prod^{N_c}_{i=1} \Phi(-\frac{\mu_i(\bm{x})}{\sigma_i(\bm{x})}).
\end{equation}
By maximizing the probability of feasibility, we favor the region that is more likely to satisfy the constraints and reduce sampling around the invalid area. However, the sequential decision-making nature of the PF function prevents it from being parallelized.
Also, the value of the PF function is easy to be zero even if there is only one constraint that violates the design specification, which significantly reduces its ability to prioritize candidate data points in practice. It is noteworthy that this problem deteriorates when the number of constraints increases.

To address this problem, we introduce two additional penalization terms to better prioritize data points and parallelize the optimization procedure. Intuitively, to reduce the amount of constraint violation, 
we can simply minimize the constraint that has a predictive mean higher than zero, \ie
\begin{equation}
    \label{eq:intuitive}
    \text{minimize} \quad \sum^{N_c}_{i=1} \text{max}(0, \mu_i(\bm{x})).
\end{equation}
However, this fitness measurement relies solely on the predictive mean and tends to be overconfident. 
For the region with high uncertainty estimation, the provided predictive mean is less reliable. For the region with low predictive uncertainty, the surrogate model has much higher confidence in its prediction and the corresponding predictive mean is more trustworthy. This means that treating constraints equally is not cost-efficient and we should pay more attention to the constraint that has a higher confidence measurement.
In other words, the amount of effort we should spend on making an arbitrary constraint feasible is negatively correlated with the uncertainty estimation. 
Therefore, we refine equation (\ref{eq:intuitive}) and scale the predictive mean of each constraint with its estimated confidence measurement. The corresponding formulation is as follow:
\begin{equation}
    \label{eq:second_term}
    \text{minimize} \quad \sum^{N_c}_{i=1} \text{max}(0, \frac{\mu_i(\bm{x})}{\sigma_i(\bm{x})}).
\end{equation}
With this adaptive measurement of the constraint violation, we focus on optimizing the constraint that violates the design specification and greatly speed up the process of searching for the first feasible point. 
However, for the region with predictive mean below zero, the adaptive constraint violation measurement can no longer handle the constraints.


To compensate disadvantages of the PF function and the adaptive constraint violation estimation, we select both PF and equation (\ref{eq:second_term}) to guide the search. In this way, the adaptive constraint violation measurement can help to prioritize data points when the PF function value is zero. The PF function can help to prioritize data points when the predictive means of all constraints satisfy design specifications and the expected constraint violation value is zero. To achieve a more extensive coverage over the Pareto front of exploration and exploitation, we also introduce the naive constraint violation measurement in equation (\ref{eq:intuitive}) to encourage exploration. In other words, we select query points by sampling $B$ data points from the Pareto front of the following equation:
\begin{equation}
    \label{eq:first_stage}
    \text{minimize} \quad -\text{PF}(\bm{x}), \sum^{N_c}_{i=1} \text{max}(0, \mu_i(\bm{x})), \sum^{N_c}_{i=1} \text{max}(0, \frac{\mu_i(\bm{x})}{\sigma_i(\bm{x})}).
\end{equation}
Thus, we can not only increase the information gain about the feasible region for each selected data but also parallelize the optimization procedure.

\vspace{-0.1in}
\subsection{Search for the Global Optimum} \label{sec:stage_two}

After finding out the first feasible point, we will try to search the state space by penalizing around the region with a high probability of violating the constraints and minimize the objective function by balancing between the exploration and exploitation. 

The state-of-the-art constrained Bayesian optimization algorithm WEIBO \cite{lyu2017efficient} handles the design specifications by weighting the expected improvement function with the probability of feasibility. By penalizing around the region that has a much higher probability to violate the constraints, the weighted expected improvement (wEI) \cite{gelbart2014bayesian, snoek2013bayesian} function favors the points that maximize the objective function while still satisfying the constraints. 

However, WEIBO is designed to work in sequential mode, which greatly limits its efficiency. To cope with batch constrained Bayesian optimization, we build upon the spirit of wEI by combining our acquisition function ensemble in equation (\ref{eq:MACE}) with the previously designed penalization strategies in equation (\ref{eq:first_stage}) to reduce sampling around the infeasible region.
The corresponding formulation is as follow:
\begin{equation}
    \label{eq:second_stage}
    \begin{aligned}
    \text{minimize} \quad &  \text{LCB}(\bm{x}), -\text{PI}(\bm{x}), -\text{EI}(\bm{x}), -\text{PF}(\bm{x}), \\
    & \sum_{i=1}^{N_c} \text{max}(0, \mu_i(\bm{x})), \sum_{i=1}^{N_c} \text{max}(0, \frac{\mu_i(\bm{x})}{\sigma_i(\bm{x})}).
    \end{aligned}
\end{equation}
Also, to reduce sampling around the region that is more likely to violate the constraints, we propose a recommendation pruning strategy and select query points that satisfy: 
\begin{equation}
    \label{eq:pruning}
    \sum_{i=1}^{N_c} \text{max}(0, \frac{\mu_i(\bm{x})}{\sigma_i(\bm{x})}) <= \rho.
\end{equation}
where $\rho$ is a user-defined parameter and we fixed it as 0.05 in this paper. With this specially designed recommendation pruning strategy, we favor the region with high probability to be valid while achieving a better tradeoff between exploration and exploitation. In this way, we can prevent the acquisition function from spending too much effort around the region that is likely to violate the constraints.








\subsection{Summary} \label{sec:summary}

The overall framework of our proposed constrained batch Bayesian optimization algorithm is presented in Algorithm \ref{algo:constrained_MACE}. By dividing the optimization procedure into two stages, we first focus on searching for the first feasible point before taking both constraints and objective into consideration. After obtaining the first feasible point, the updated surrogate model provides more informative posterior distribution about whether an arbitrary design satisfies constraints. Therefore, the overall time consumption of the optimization process can be significantly reduced. 
By adopting a new penalization scheme, we guide the decision-making process in parallel by favoring the potential area with better objective function value and penalizing around the region that is likely to violate the constraints. Thanks to the above, the global optimum that satisfies constraints can be reached after a limited number of iterations.

\begin{algorithm}
    \caption{Constrained MACE algorithm}
    \label{algo:constrained_MACE}
    \begin{algorithmic}[1]
        \Require The size of the initial dataset $N_{init}$, the maximum number of iteration $N_{iter}$, and the batch size $B$.
        \State Randomly sample an initial dataset $D_0=\{X, \bm{y}\}$
        \For{$t = 0 \to N_{iter}$}
        \State Construct Gaussian process regression model with training dataset $D_t$
        \If{The first feasible point has been achieved} {Generate a Pareto-optimal dataset with equation (\ref{eq:second_stage}), and pruning the candidate pool with equation (\ref{eq:pruning}) to get $P_t$}
        \Else{} {Generate a Pareto-optimal dataset $P_t$ with equation (\ref{eq:first_stage})}
        \EndIf
        \State Randomly sample $B$ data points $X_t = \{\bm{x}_{t,1}, \cdots, \bm{x}_{t,B}\}$ from $P_t$
        \State Evaluate the sampled data points $\bm{y}_t = \{f(\bm{x}_{t,1}), \cdots, f(\bm{x}_{t,B})\}$
        \State $D_{t+1} = \{D_t, \{X_t, \bm{y}_t\}\}$
        \EndFor
        \State \Return The best optimization result that satisifies the constraints.
    \end{algorithmic}
\end{algorithm}

\vspace{-0.2in}
\section{Experimental Results} \label{sec:experiment}

In this section, we experimentally evaluate the performances of MACE on four real-world analog circuits to show the benefit of our batch design. We examine the efficiency and effectiveness of MACE on both constrained and unconstrained optimization problems and quantitatively compare its performances with the state-of-the-art optimization algorithms. 
Our proposed MACE algorithm is implemented in Python with GPy \cite{gpy2014} and Platypus\footnote{https://github.com/Project-Platypus/Platypus} libraries. 
All circuit performances are generated with commercial HSPICE circuit simulator. 
All experiments are conducted on a Linux workstation with two Intel Xeon CPUs and 128GB memory.


For the unconstrained optimization problem, we first run the experiments in sequential mode to demonstrate that our acquisition function design can achieve a better tradeoff between exploration and exploitation. We also evaluate the performances of MACE in three different batch sizes (B=5, 10, 15) to explore the impact of batch size and shows the effectiveness of our proposed batch design. To ensure a fair comparison, we run each Bayesian optimization algorithms with the same simulation budget, regardless of the batch size. To give a quantitative measurement of the cost-effectiveness and stability of each algorithm, we run each algorithm 20 times to reduce the random fluctuations and present the optimization results in terms of the best-case, worst-case, mean, and standard deviation. To differentiate between the sequential and batch mode, we label different methods using batch policy type followed by the batch size.

In \textit{sequential} mode, we compare the performances of MACE with 3 state-of-the-art optimization algorithms: 
(1) DE \cite{liu2009analog}, which is an optimization strategy based on differential evolution. 
(2) EI \cite{mockus1978application}, which is an improvement-based acquisition function for Bayesian optimization framework that guides the search by maximizing the expected improvement. 
(3) LCB \cite{srinivas2009gaussian}, which is an optimistic strategy that traverses the design space with theoretical cumulative regret bound.

In \textit{batch} mode, MACE is compared with 6 state-of-the-art batch policies based on the Bayesian optimization framework:
(1) pBO \cite{hu2018parallelizable}, which selects the batch by introducing weighting parameter to balance between the exploration and exploitation.
(2) pHCBO \cite{hu2018parallelizable}, which encourages diversity in a batch by introducing a well-designed penalization scheme.
(3) LP-EI \cite{gonzalez2016batch}, which explores the design space with EI and maintains diversity in a batch with local penalization term.
(4) LP-LCB \cite{gonzalez2016batch}, which guides the search with LCB and introduces a local repulsive term to penalize around the early decisions in a batch.
(5) BLCB \cite{desautels2014parallelizing}, which reduces redundantly sampling around busy locations with hallucinated observations.
(6) GPUCB-PE \cite{Contal2013Parallel}, which combines the benefits of UCB policies with pure exploration queries in the same batch to improve information gain per observation.

To further investigate the relative merits of different acquisition function ensembles, we also compare MACE with 3 additional batch alternatives in both \textit{sequential} and \textit{batch} mode:
(1) PI-EI, which selects the batch from the Pareto front of PI and EI.
(2) EI-LCB, which explores the design space with EI and LCB ensemble.
(3) PI-LCB, which combines the benefits of PI and LCB to guide the search.

For the constrained optimization problem, we compare the performances of MACE with 6 well-designed optimization algorithms: 
(1) WEIBO \cite{lyu2017efficient}, is a Bayesian optimization algorithm based on the weighted expected improvement function.
(2) GASPAD \cite{liu2014gaspad:}, is a surrogate mode-aware evolutionary search algorithm that favors the valid region with the selection-based constraint handling method.
(3) MSP \cite{yang2018smart-msp:}, is a self-adaptive multiple starting point approach that tries to approximate the global optimum by learning from the previous local search.
(4) DE \cite{liu2009analog}, is an optimization algorithm based on the differential evolutionary methodology.
(5) PSO \cite{vural2012analog}, is the particle swarm optimization methodology that tries to mimic the biological process to obtain the global optimum.
(6) SA \cite{phelps2000anaconda:},  is the simulated annealing algorithm that guides the search by simulating the physical process.
To further investigate the impact of the two-stage approach, we also compare the performance of MACE with its one-stage counterpart \textit{o}MACE, which only selects the batch with the second stage design of MACE framework.

To ensure a fair comparison, we respectively run each algorithm 12 times to average the random fluctuations and present the number of runs that satisfy the constraints. To quantitatively evaluate the performances of each algorithm, we present the optimization results in terms of the mean, median, best-case and worst-case results. For simplicity of comparison, we record the equivalent simulation time consumption with the equivalent number of circuit simulations on average to achieve the final circuit design (Avg. \# Sim). We also record the number of runs that successfully find feasible designs for each algorithm (\# Success). For each algorithm, we also present the constraint function values of the best design in all runs.
\subsection{Two-Stage Operational Amplifier}


\begin{figure}
	\centering
	\includegraphics[width=0.38\textwidth]{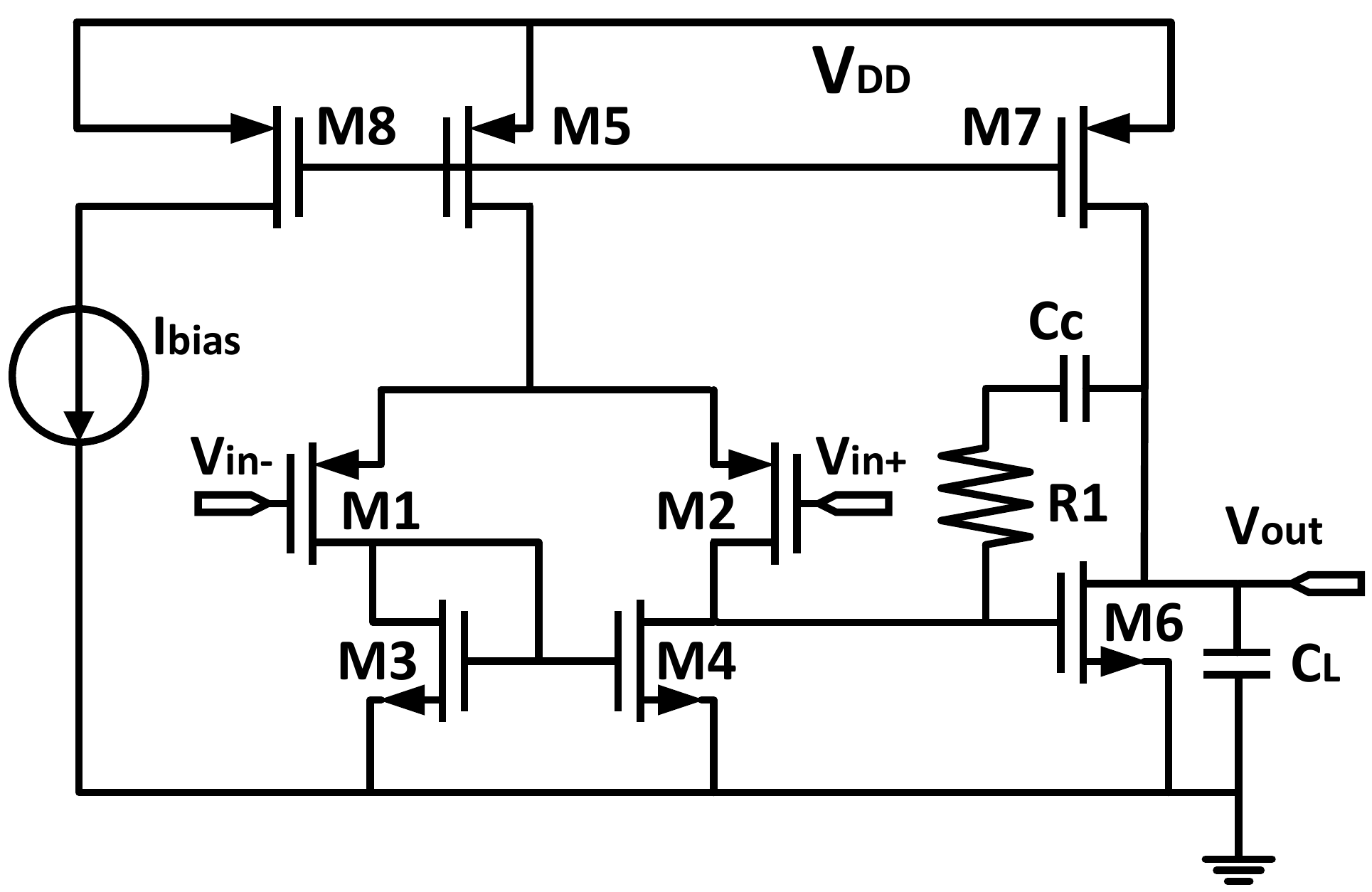}
	\vspace{-0.1in}
	\caption{The schematic of the two-stage operational amplifier circuit, which is reproduced from \cite{yang2018smart-msp:}.\vspace{-3mm}}
	\label{fig:two_amplifier}
\end{figure}

The schematic of the two-stage operational amplifier circuit is presented in Figure \ref{fig:two_amplifier}. As proposed in \cite{yang2018smart-msp:}, this circuit is implemented in a SMIC 180nm process and has a total of 10 design variables, including lengths and widths of the transistors, the resistance of the resistors, and the capacitance of the capacitors. In this circuit, we seek to maximize the open-loop gain (GAIN), the unity gain frequency (UGF), and the phase margin (PM) at the same time. We evaluate the performances of MACE on both unconstrained and constrained optimization problems. 

\textbf{Unconstrained optimization.} For the unconstrained optimization problem, our formulated design specification is as follow:
\begin{equation}
	\text{maximize} \quad 1.2 \times GAIN + 10.0 \times UGF + 1.6 \times PM.
\end{equation}

To fully tap the potential of MACE, we conduct experiments in both sequential and batch mode. 
In batch mode, we run each algorithm with 3 different batch sizes (B=5, 10, 15) to demonstrate the impact of different batch sizes. For algorithms in Bayesian optimization literature, we set the size of the initial dataset as 20 and limit the maximum number of simulations as 270, regardless of the batch size. For DE, the simulation budget is set to be 20000. To ensure a fair comparison, we repetitively run each algorithm 20 times to eliminate the random fluctuations. The unconstrained optimization results are presented in Table \ref{tb:two_amplifier1}.

\begin{table}
    \centering
    \caption{The unconstrained optimization results of the two-stage operational amplifier circuit.}
    \label{tb:two_amplifier1}
    \begin{tabular}{ccccc}
	\hline
	Algo & Best & Worst & Mean & Std \\
	\hline
	DE & 685.44 & 680.65 & 682.19 & 1.56 \\
    LCB & 690.35 & 685.02 & 688.68 & 1.51 \\
    EI & 690.29 & 670.49 & 688.07 & 4.98 \\
    PI-EI & 690.35 & 690.27 & 690.34 & 0.03 \\
    EI-LCB & 690.35 & 690.27 & 690.34 & 0.03 \\
    PI-LCB & 690.35 & 688.09 & 690.23 & 0.49 \\
	MACE & 690.36 & 690.27 & \textbf{690.34} & 0.03 \\
	\hline
    pBO-5 & 690.36 & 688.00 & 689.59 & 1.01 \\
    pHCBO-5 & 690.36 & 615.27 & 678.47 & 20.82 \\
    LP-EI-5 & 690.15 & 664.68 & 685.15 & 8.34 \\
    LP-LCB-5 & 690.18 & 660.42 & 685.25 & 8.87 \\
    BLCB-5 & 690.33 & 685.70 & 688.69 & 1.28 \\
    GPUCB-PE-5 & 690.35 & 618.41 & 679.33 & 23.52 \\
    PI-EI-5 & 690.35 & 653.74 & 687.72 & 9.42 \\
    EI-LCB-5 & 690.35 & 680.24 & 689.35 & 2.56 \\
    PI-LCB-5 & 690.35 & 689.85 & 690.28 & 0.13 \\
	MACE-5 & 690.36 & 690.27 & \textbf{690.33} & 0.03 \\
	\hline
    pBO-10 & 690.36 & 535.80 & 676.45 & 39.49 \\
    pHCBO-10 & 690.36 & 641.83 & 685.67 & 12.24 \\
    LP-EI-10 & 689.93 & 633.32 & 677.67 & 17.25 \\
    LP-LCB-10 & 690.12 & 663.58 & 685.37 & 6.51 \\
    BLCB-10 & 690.36 & 685.09 & 688.77 & 1.47 \\
    GPUCB-PE-10 & 690.28 & 575.75 & 658.14 & 37.18 \\
    PI-EI-10 & 690.35 & 655.14 & 684.33 & 10.91 \\
    EI-LCB-10 & 690.27 & 673.54 & 686.31 & 4.98 \\
    PI-LCB-10 & 690.35 & 533.55 & 682.30 & 34.13 \\
	MACE-10 & 690.36 & 685.45 & \textbf{690.00} & 1.26 \\
	\hline
    pBO-15 & 690.35 & 558.18 & 673.52 & 34.15 \\
    pHCBO-15 & 690.36 & 608.29 & 681.25 & 22.71 \\
    LP-EI-15 & 689.70 & 609.82 & 677.53 & 20.54 \\
    LP-LCB-15 & 687.88 & 641.51 & 668.71 & 15.77 \\
    BLCB-15 & 690.36 & 685.78 & 688.73 & 1.36 \\
    GPUCB-PE-15 & 689.35 & 612.63 & 672.49 & 22.15 \\
    PI-EI-15 & 690.32 & 500.57 & 659.80 & 56.18 \\
    EI-LCB-15 & 690.07 & 654.13 & 678.93 & 11.55 \\
    PI-LCB-15 & 690.35 & 522.36 & 681.03 & 36.45 \\
	MACE-15 & 690.35 & 682.97 & \textbf{688.96} & 2.28 \\
	\hline
    \end{tabular}
    \vspace{-0.15in}
\end{table}

\begin{table*}
    \centering
    \caption{The constrained optimization results of the two-stage operational amplifier circuit.}
    \label{tb:two_amplifier2}
    \begin{tabular}{ccccccccccccc}
        \hline
        Algo & \textit{o}MACE-5 & MACE-5 & \textit{o}MACE-10 & MACE-10 & \textit{o}MACE-15 & MACE-15 & WEIBO & GASPAD & MSP & DE & PSO & SA \\
        \hline
        UGF/MHz & 40.00 & 40.01 & 40.00 & 40.00 & 40.02 & 40.03 & 40.03 & 40.20 & 40.01 & 40.10 & fail & fail \\
        PM/$^{o}$ & 61.03 & 61.03 & 60.98 & 60.99 & 60.94 & 60.93 & 60.87 & 60.83 & 61.43 & 60.95 & fail & fail \\
        \hline
        GAIN(mean) & 90.05 & \textbf{90.06} & 89.93 & 89.93 & 89.75 & 89.94 & 89.61 & 89.24 & 89.42 & 89.42 & fail & fail \\
        GAIN(median) & 90.05 & \textbf{90.11} & 89.96 & 90.00 & 89.96 & 90.07 & 89.73 & 89.24 & 89.35 & 89.38 & fail & fail \\
        GAIN(best) & 90.18 & \textbf{90.18} & 90.14 & 90.15 & 90.16 & 90.18 & 90.15 & 89.95 & 89.95 & 89.67 & fail & fail \\
        GAIN(worst) & \textbf{89.89} & 89.84 & 89.55 & 89.57 & 88.61 & 89.81 & 87.71 & 88.50 & 89.00 & 89.09 & fail & fail \\
        \hline
        Avg. \# Sim & 103 & 74 & 53 & 39 & 36 & 32 & 170 & 385 & 5263 & 5931 & fail & fail \\
        \# Success & 12/12 & 12/12 & 12/12 & 12/12 & 12/12 & 12/12 & 12/12 & 12/12 & 12/12 & 12/12 & 0/12 & 0/12 \\
        \hline
    \end{tabular}
    \vspace{-0.15in}
\end{table*}

We start with examining the experimental results in \textit{sequential} mode, which are presented in the top block of Table \ref{tb:two_amplifier1}. Clearly, MACE not only outperforms the state-of-the-art optimization algorithms on average, but also have a much smaller deviation. Compared with DE, our proposed algorithm reduces the simulation time by $74\times$ while obtaining better optimization results. Compared with EI, LCB, PI-EI, EI-LCB and PI-LCB, MACE demonstrates a much higher convergence rate and more stable performance. This clearly shows that MACE can better guide the search by simultaneously maximizing the immediate improvement value and the cumulative regret of each candidate point in sequential mode. Another interesting observation is that the acquisition function ensembles (PI-EI, EI-LCB, PI-LCB and MACE) always obtain better optimization results compared to EI and LCB, which rely solely on a single acquisition function to search the state space. This reveals that the acquisition function ensemble can combine the benefits of several acquisition functions and increase the information gain per data point.

Now we move on to analyze the performances of MACE in \textit{batch} mode. As expected, MACE consistently outperforms the state-of-the-art batch policies for the same batch size, which demonstrates both the efficiency and effectiveness of our batch design. Compared with pBO, pHCBO, LP-EI, LP-LCB, BLCB and GPUCB-PE that select data points iteratively and greedily for each batch, MACE simply constructs the batch at once by randomly selecting query points from the Pareto front of PI, EI and LCB. In this way, MACE combines the strengths of several state-of-the-art acquisition functions. Instead of designing the penalization scheme manually and penalizing around the early observations in a batch with human biased presumptions, MACE naturally maintains diversity within the same batch. The optimization results of pBO and pHCBO, when the batch size is 5, further reveal that manually design penalization scheme sometimes can even hinder the optimization process. Compared with PI-EI, EI-LCB and PI-LCB that guide the search with only two acquisition functions, MACE consistently achieves better optimization results across different batch sizes. This indicates that different acquisition functions have different characteristics in selecting query points, and the efficiency of the acquisition function ensemble improves with the increasing number of selected acquisition functions. The comparably small deviations across different batch sizes also demonstrate the stability and robustness of MACE.


To further analyze the impact of batch size, we compare the optimization results across different batch sizes. From an information gain perspective, the performances in sequential mode can always be seen as a baseline for the batch policies with the same acquisition function for decision selections. The optimization results of EI, LCB, PI-EI, EI-LCB and PI-LCB in sequential mode are consistently better compared to its batch counterparts. This is because the two-stage operational amplifier circuit requires more exploitation than exploration to optimization. Compared with all the other batch policies the optimization results of which deteriorate quickly with the increase of the batch size, MACE demonstrates strong stability and robustness in terms of both the average results and the standard deviations. For BLCB, despite getting similar results for different batch sizes, its consistently outperformed optimization results of BLCB reveal that searching the design space by solely relying on a single utility function can greatly limit the efficiency of decision making. The experimental results of the unconstrained optimization problem provide quantitative evidence for the efficiency and robustness of our batch design. Another noteworthy phenomenon is that with a fixed number of selected acquisition functions, different acquisition function combinations have different behavior patterns. With the increase of the batch size, the performance of PI-LCB deteriorates much slower than PI-EI and EI-LCB. This is due to the fact that PI tends to do more exploitation, LCB tends to do more exploration and EI stays relatively in the middle. Thus, PI-LCB naturally has better coverage over the exploitation-exploration Pareto front than PI-EI and EI-LCB. With the increase of the batch size, PI-EI and EI-LCB become too conservative and PI-LCB starts to outperform them.



\textbf{Constrained optimization.} For the constrained optimization problem, our formulated design specification is as follow:
\begin{equation}
	\begin{aligned}
		\text{maximize} \quad & GAIN & & \\
		\text{s.t.} \quad & UGF \quad & > \quad & 40MHz, \\
		\quad & PM \quad & > \quad & 60^o.
	\end{aligned}
\end{equation}

In this experiment, we compare the performances of MACE and \textit{o}MACE with the state-of-the-art constrained optimization algorithms, including WEIBO, GASPAD, MSP, DE, PSO and SA. To fully explore the potential of our constrained batch design, we test both MACE and \textit{o}MACE in 3 different batch sizes (B=5, 10, 15) and fix the maximum number of simulations as 620, regardless of the batch size. For MACE and \textit{o}MACE, the size of the initial dataset is 20. For WEIBO, we randomly sample 20 initial data point and set the simulation budget as 200. For GASPAD, we limit the maximum number of simulations as 500. As for the rest of the algorithms, we limit the simulation budget as 10000. To reduce the random fluctuations and fairly compare the experimental results, we repetitively run each algorithm 12 times. The corresponding optimization results are presented in Table \ref{tb:two_amplifier2}.

It is worth notice that the optimization results of both PSO and SA fail to meet the constraints, which means the design specification is hard to satisfy. Compared with DE, MSP, GASPAD and WEIBO, MACE respectively reduces the simulation time by up to $185\times$, $164\times$, $12\times$ and $5\times$, while obtaining relatively better optimization results. The experimental results quantitatively demonstrate that the optimization process can be significantly sped up by proposing a batch of data points at each iteration and assigning the circuit simulation to different workers. Also, the fact that MACE consistently outperforms its one-stage counterpart regardless of the batch size shows that the two-stage policy can greatly reduce sampling around the infeasible region, thus, accelerate the optimization procedure. We further investigate the impact of different batch sizes. Although the performance of MACE deteriorates slightly with the increase of the batch size, MACE still obtains much better optimization results in terms of mean, median, best-case and worst-case results when the batch size 15. This clearly demonstrates the robustness and stability of our proposed algorithm.



\subsection{Class-E Power Amplifier}


\begin{figure}
	\centering
	\includegraphics[width=0.40\textwidth]{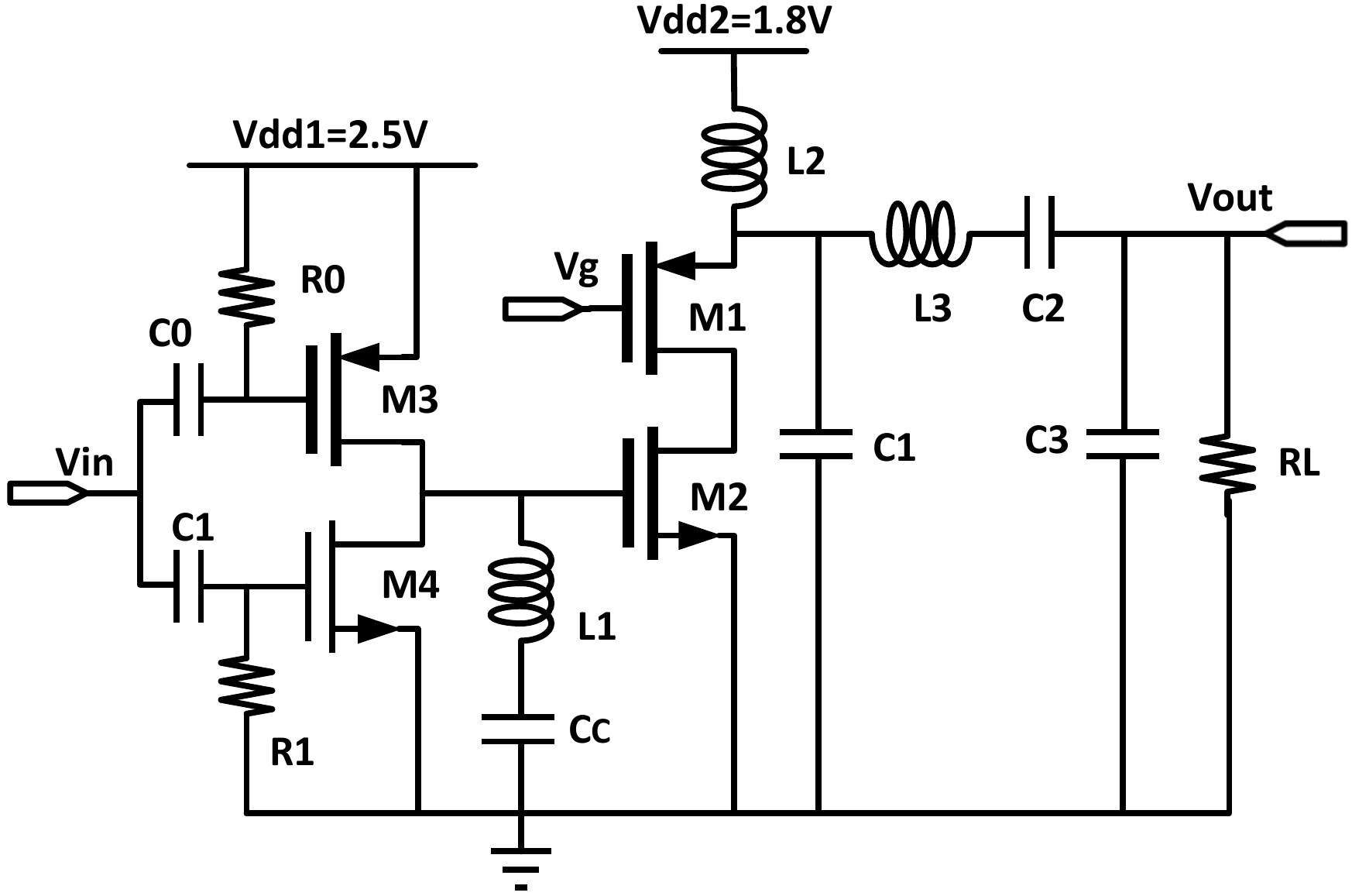}
	\vspace{-0.1in}
	\caption{The schematic of the class-E power amplifier circuit, which is reproduced from \cite{lyu2018batch}.\vspace{-0.1in}}
	\label{fig:class-E}
\end{figure}

The second circuit we evaluate MACE with is the class-E power amplifier circuit, which is shown in Figure \ref{fig:class-E}. Implemented in a 180nm TSMC process, this circuit has a total of 12 design variables. The corresponding circuit performances are generated via the commercial HSPICE circuit simulator. For this circuit, our target of design is to maximize the power-added efficiency (PAE) and the output power (Pout) simultaneously. 

\textbf{Unconstrained optimization.} Our formulated design specification for the unconstrained optimization problem is:
\begin{equation}
    \text{maximize} \quad 3 \times PAE + Pout.
    \label{eq:class-E1}
\end{equation}

To ensure a fair comparison, we run each algorithm 20 times to average the random fluctuations. For DE, we set the maximum number of simulations as 15000. For the rest of the Bayesian optimization algorithms, we compare the performances with a fixed simulation budget across different batch sizes to fully explore the potential of our proposed algorithm. Specifically, we set the size of the initial dataset as 20 and limit the maximum number of simulations as 470. The corresponding unconstrained optimization results are presented in Table \ref{tb:class-E1}.

\begin{table} 
    \centering
    \caption{The unconstrained optimization results of the class-E power amplifier circuit.}
    \label{tb:class-E1}
    \begin{tabular}{ccccc}
        \hline
        Algo & Best & Worst & Mean & Std \\
        \hline
	    DE & 4.56 & 4.33 & 4.43 & 0.08 \\
        LCB & 4.10 & 3.59 & 3.89 & 0.14 \\
        EI & 4.13 & 3.52 & 3.85 & 0.19 \\
        PI-EI & 5.80 & 4.20 & 4.63 & 0.34 \\
        EI-LCB & 5.11 & 4.09 & 4.45 & 0.25 \\
        PI-LCB & 4.83 & 3.99 & 4.49 & 0.26 \\
	    MACE & 6.12 & 3.85 & \textbf{4.79} & 0.57 \\
	    \hline
        pBO-5 & 4.61 & 3.76 & 4.17 & 0.19 \\
        pHCBO-5 & 4.42 & 3.66 & 4.16 & 0.16 \\
        LP-EI-5 & 4.17 & 3.51 & 3.87 & 0.19 \\
        LP-LCB-5 & 4.26 & 3.43 & 3.82 & 0.26 \\
        BLCB-5 & 4.73 & 4.07 & 4.23 & 0.14 \\
        GPUCB-PE-5 & 4.51 & 3.62 & 4.17 & 0.19 \\
        PI-EI-5 & 4.60 & 3.55 & 4.22 & 0.27 \\
        EI-LCB-5 & 5.88 & 3.79 & 4.43 & 0.51 \\
        PI-LCB-5 & 4.98 & 3.65 & 4.49 & 0.31 \\
	    MACE-5 & 4.83 & 3.76 & \textbf{4.51} & 0.31 \\
        \hline
        pBO-10 & 4.34 & 3.80 & 4.11 & 0.16 \\
        pHCBO-10 & 4.82 & 3.79 & 4.17 & 0.23 \\
        LP-EI-10 & 4.60 & 3.49 & 3.86 & 0.28 \\
        LP-LCB-10 & 4.26 & 3.59 & 3.88 & 0.20 \\
        BLCB-10 & 4.46 & 3.94 & 4.26 & 0.13 \\
        GPUCB-PE-10 & 4.40 & 3.66 & 4.04 & 0.21 \\
        PI-EI-10 & 4.82 & 2.85 & 4.13 & 0.52 \\
        EI-LCB-10 & 4.97 & 3.39 & 4.29 & 0.38 \\
        PI-LCB-10 & 5.65 & 3.61 & 4.46 & 0.45 \\
	    MACE-10 & 5.54 & 3.73 & \textbf{4.64} & 0.46 \\
        \hline
        pBO-15 & 4.61 & 3.87 & 4.17 & 0.19 \\
        pHCBO-15 & 4.31 & 3.67 & 4.10 & 0.16 \\
        LP-EI-15 & 4.33 & 3.50 & 3.87 & 0.23 \\
        LP-LCB-15 & 4.28 & 3.58 & 3.85 & 0.17 \\
        BLCB-15 & 4.70 & 4.10 & 4.30 & 0.13 \\
        GPUCB-PE-15 & 4.20 & 3.54 & 3.93 & 0.17 \\
        PI-EI-15 & 4.78 & 3.06 & 4.14 & 0.43 \\
        EI-LCB-15 & 4.96 & 3.67 & 4.20 & 0.30 \\ 
        PI-LCB-15 & 4.77 & 3.31 & 4.28 & 0.37 \\
        MACE-15 & 5.41 & 3.39 & \textbf{4.32} & 0.42 \\
	\hline
    \end{tabular}
    \vspace{-0.15in}
\end{table}


\begin{table*}
    \centering
    \caption{The constrained optimization results of the class-E power amplifier circuit.}
    \label{tb:class-E2}
    \begin{tabular}{ccccccccccccc}
        \hline
        Algo & \textit{o}MACE-5 & MACE-5 & \textit{o}MACE-10 & MACE-10 & \textit{o}MACE-15 & MACE-15 & WEIBO & GASPAD & MSP & DE & PSO & SA \\
        \hline
        Pout/dBm & 2.11 & 3.36 & 2.23 & 2.73 & 3.04 & 2.94 & 2.04 & 2.21 & 2.31 & 2.21 & 2.33 & 2.01 \\
        \hline
        PAE(mean) & 0.83 & \textbf{0.85} & 0.83 & 0.84 & 0.80 & 0.83 & 0.74 & 0.78 & 0.73 & 0.73 & 0.71 & 0.66 \\
        PAE(median) & 0.80 & \textbf{0.81} & 0.79 & 0.80 & 0.79 & 0.80 & 0.75 & 0.75 & 0.71 & 0.72 & 0.70 & 0.67 \\
        PAE(best) & 0.97 & \textbf{1.06} & 0.97 & 1.10 & 0.92 & 1.03 & 0.76 & 0.95 & 0.98 & 0.77 & 0.86 & 0.72 \\
        PAE(worst) & 0.75 & \textbf{0.76} & 0.73 & 0.70 & 0.75 & 0.73 & 0.73 & 0.72 & 0.66 & 0.69 & 0.66 & 0.60 \\
        \hline
        Avg. \# Sim & 156 & 124 & 63 & 70 & 51 & 52 & 583 & 516 & 2580 & 3657 & 3610 & 2986 \\
        \# Success & 12/12 & 12/12 & 11/12 & 12/12 & 12/12 & 12/12 & 12/12 & 12/12 & 12/12 & 12/12 & 12/12 & 7/12 \\
        \hline
    \end{tabular}
    \vspace{-0.1in}
\end{table*}

We start with analyzing the performances of MACE in \textit{sequential} mode presented in the top block of Table \ref{tb:class-E1}. Compared with EI and LCB, the acquisition function ensembles achieve much better optimization results with the same simulation budget. This observation again shows that the sampling efficiency of the acquisition function ensembles is much higher than the state-of-the-art acquisition functions. By searching the design space with several acquisition functions, the acquisition function ensembles can combine the benefits of the state-of-the-art utility functions and better guide the search. The fact that MACE consistently outperforms PI-EI, EI-LCB and PI-LCB further confirms that the performance of the acquisition function ensemble improves with the increasing number of acquisition functions. Compared with DE, MACE reduces the simulation time by up to $42\times$ while obtaining more competitive optimization results. The superior performance of MACE in terms of the best-case result further shows that MACE has great potential in selecting the query point. Our results demonstrate that MACE can lead the search more efficiently and effectively.

We now move on to investigate the performances of MACE with the state-of-the-art \textit{batch} policies in the batch mode. To fully exploit the latent capacity of MACE and demonstrate the impact of different batch sizes, we run each algorithm in 3 different batch sizes (B=5, 10, 15) and compare the experimental results with the state-of-the-art batch policies. Overall, MACE consistently outperforms the state-of-the-art batch policies with respect to the same batch size. For pBO, pHCBO, LP-EI and LP-LCB, the corresponding optimization results for different batch sizes are comparably the same and sometimes even better than their sequential counterparts. This reveals that batch policy doesn't necessarily lead to penalized information gain per data point. The fact that the performances of BLCB improve with the increase of the batch size further demonstrates that a relative magnitude of batch size encourages exploration. Besides, the class-E power amplifier circuit actually requires more exploration than exploitation to fully search the design space. For GPUCB-PE, the optimization results indicate that GPUCB-PE naturally encourages more exploitation than exploration. As for MACE, the consistently competitive performances, regardless of the batch size, demonstrate that MACE can achieve a better tradeoff between exploration and exploitation. Instead of relying on a single acquisition function to guide the search, MACE selects query points from the Pareto front of the acquisition function ensemble. Therefore, MACE can always select more informative data point to better facilitate the optimization process. The experimental results presented in Table \ref{tb:class-E1} demonstrate the efficiency and effectiveness of MACE for both sequential and batch mode.


\begin{table*}
    \centering
    \caption{The constrained optimization results of the low-power three-stage amplifier circuit, the results of MSP, DE, PSO and SA come from \cite{lyu2017efficient}.}
    \label{tb:three_amplifier}
    \begin{tabular}{ccccccccccccc}
        \hline
        Algo & \textit{o}MACE-5 & MACE-5 & \textit{o}MACE-10 & MACE-10 & \textit{o}MACE-15 & MACE-15 & WEIBO & GASPAD & MSP & DE & PSO & SA \\
        \hline
        GAIN/dB & 101.36 & 102.37 & 101.25 & 102.65 & 102.58 & 101.08 & 100.67 & 100.82 & 100.81 & 102.73 & 102.39 & 102.49 \\
        UGF/MHz & 0.92 & 0.92 & 0.92 & 0.92 & 0.93 & 0.92 & 0.93 & 0.94 & 0.98 & 0.96 & 0.96 & 1.05 \\
        PM/$^{o}$ & 52.52 & 52.50 & 52.51 & 52.70 & 53.50 & 52.51 & 53.10 & 52.66 & 53.22 & 54.62 & 54.25 & 56.70 \\
        GM/dB & 19.80 & 19.53 & 19.86 & 19.50 & 19.55 & 19.50 & 19.58 & 19.83 & 22.30 & 20.62 & 21.32 & 20.92 \\
        SR+(V/$\mu$s) & 0.21 & 0.20 & 0.20 & 0.20  & 0.24 & 0.20 & 0.19 & 0.21 & 0.23 & 0.21 & 0.23 & 0.25 \\
        SR-(V/$\mu$s) & 0.41 & 0.37 & 0.51 & 0.43 & 0.37 & 0.48 & 0.41 & 0.54 & 0.51 & 0.54 & 0.51 & 0.49 \\
        \hline
        Iq(mean) & 31.67 & 30.61 & 33.01 & 30.67 & 33.10 & \textbf{29.91} & 37.78 & 35.08 & 49.60 & 41.26 & 44.22 & 59.78 \\
        Iq(median) & 30.87 & 29.83 & 32.41 & \textbf{29.75} & 33.38 & 30.24 & 34.90 & 35.64 & 48.29 & 40.70 & 43.16 & 53.56 \\
        Iq(best) & 29.34 & 27.27 & 28.26 & 27.18 & 29.24 & \textbf{26.54} & 29.28 & 29.26 & 32.06 & 37.32 & 37.18 & 46.34 \\
        Iq(worst) & 36.74 & 35.65 & 38.22 & 34.98 & 36.59 & \textbf{32.80} & 49.89 & 38.77 & 75.32 & 46.09 & 59.56 & 83.48 \\
        \hline
        Avg. \# Sim & 166 & 172 & 85 & 80 & 58 & 57 & 396 & 743 & 2163 & 2400 & 2417 & 620 \\
        \# Success & 12/12 & 12/12 & 12/12 & 12/12 & 12/12 & 12/12 & 12/12 & 12/12 & 9/12 & 12/12 & 12/12 & 5/12 \\
        \hline
    \end{tabular}
    \vspace{-0.1in}
\end{table*}

\textbf{Constrained optimization.} Our formulated design specification for the constrained optimization problem is:
\begin{equation}
    \begin{aligned}
        \text{maximize} \quad & PAE & & \\
        \text{s.t.} \quad & Pout \quad & > \quad & 2.0 dBm, \\ 
    \end{aligned}
    \label{eq:class-E2}
\end{equation}

In this experiment, we test both MACE and \textit{o}MACE with a fixed simulation budget for different batch sizes to explore the impact of batch size. For MACE, \textit{o}MACE and WEIBO, we randomly sample 20 initial data points and set the maximum number of simulations as 920. For GASPAD, the simulation budget is limited to 600. For the rest of the algorithms, the maximum number of simulations is 5000. To average the random fluctuations, we run each algorithm 12 times and present the mean, median, best-case and worst-case results. The corresponding constrained optimization results are presented in Table \ref{tb:class-E2}.


As expected, MACE achieves much better optimization results than the state-of-the-art optimization algorithms. 
The fact that SA and \textit{o}MACE fail to successfully find a feasible design in all runs shows that the design specification is hard to satisfy. Considering that MACE successfully obtains valid designs in all runs and consistently outperforms \textit{o}MACE, the two-stage approach can greatly help to relieve the burden of searching for the feasible region. Compared with WEIBO, GASPAD, MSP, DE and PSO, MACE reduces the simulation time by up to $11\times$, $9\times$, $49\times$, $70\times$ and $69\times$ respectively. Considering that GASPAD achieves better optimization results than WEIBO while requiring less simulation time on average, the response surface of the design specification can be well approximated with a relatively small number of data points. This also reveals that the constrained optimization of the class-E power amplifier circuit requires more exploitation than exploration. Another noteworthy phenomenon is that MACE exhibits much higher sampling efficiency compared to WEIBO when the batch size is 15. Considering that both MACE and WEIBO are assigned with the same amount of simulation budget, this suggests that our proposed batch policy can greatly accelerate the optimization process without penalization of the information gain per data points.

\subsection{Low-Power Three-Stage Amplifier}

\begin{figure}
	\centering
	\includegraphics[width=0.45\textwidth]{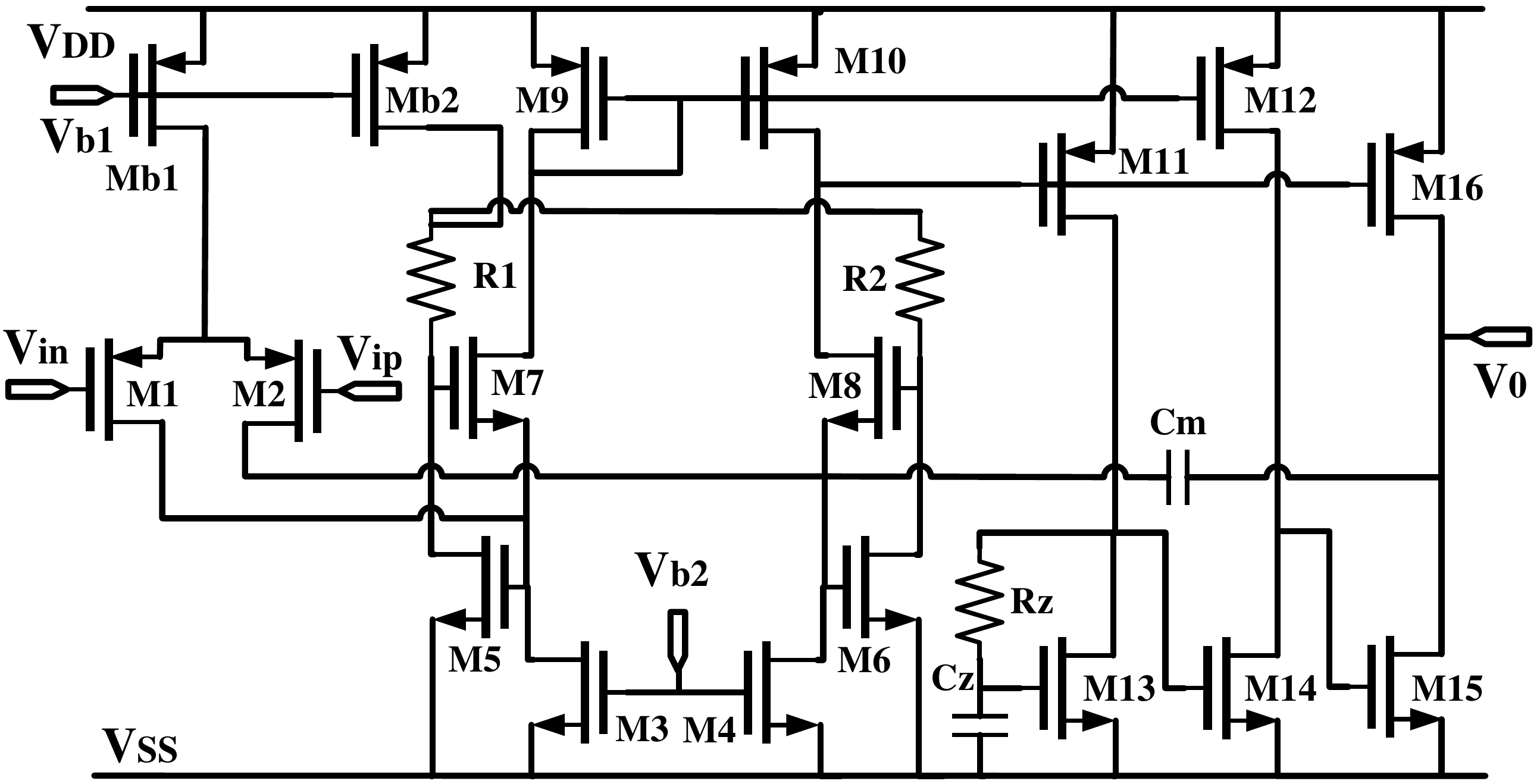}
	\caption{The schematic of the low-power three-stage amplifier circuit, which is reproduced from \cite{lyu2017efficient}.\vspace{-0.1in}}
	\label{fig:three_amplifier}
\end{figure}

The third circuit for testing is the low-power three-stage amplifier circuit. The corresponding schematic is presented in Figure \ref{fig:three_amplifier}, which is proposed in \cite{6407150}. In this circuit, there are a total of 24 design variables, including the lengths and widths of the transistors, the capacitance and resistance and the bias current. For this circuit, we only run experiments on the constrained optimization problem.

\textbf{Constrained optimization.} Our formulated constrained design specification is:
\begin{equation}
	\begin{aligned}
		\text{minimize} \quad & Iq & & \\
		\text{s.t.} \quad & GAIN \quad & > \quad & 100\text{dB}, \\
		            \quad & UGF \quad & > \quad & 0.92\text{MHz}, \\
		            \quad & PM \quad & > \quad & 52.5^{o}, \\
		            \quad & GM \quad & > \quad & 19.5\text{dB}, \\
		            \quad & SRR \quad & > \quad & 0.18\text{V/}\mu \text{s}, \\
		            \quad & SRF \quad & > \quad & 0.2\text{V/}\mu \text{s}, \\
	\end{aligned}
\end{equation}
where Iq is the static current, GAIN denotes the DC gain, UGF represents the unit gain frequency, PM stands for the phrase margin, GM means gain margin, SRR and SRF refers to the rising and falling slew rate.

\begin{table*}
    \centering
    \caption{The constrained optimization results of the charge pump circuit, the results of WEIBO, GASPAD, MSP, DE, PSO and SA come from \cite{lyu2017efficient}.}
    \label{tb:charge_pump}
    \begin{tabular}{ccccccccccccc}
        \hline
        Algo & \textit{o}MACE-5 & MACE-5 & \textit{o}MACE-10 & MACE-10 & \textit{o}MACE-15 & MACE-15 & WEIBO & GASPAD & MSP & DE & PSO & SA \\
        \hline
        diff1 & 5.43 & 5.59 & 5.56 & 5.50 & 5.98 & 5.86 & 6.58 & 6.83 & 17.81 & 17.97 & fail & fail \\
        diff2 & 4.75 & 4.44 & 4.50 & 4.48 & 5.10 & 4.45 & 5.30 & 5.28 & 16.82 & 15.49 & fail & fail \\
        diff3 & 0.06 & 0.15 & 0.12 & 0.06 & 0.10 & 0.14 & 0.24 & 0.29 & 1.51 & 1.84 & fail & fail \\
        diff4 & 0.06 & 0.22 & 0.24 & 0.07 & 0.12 & 0.18 & 0.37 & 0.40 & 2.57 & 3.56 & fail & fail \\
        deviation & 0.39 & 0.20 & 0.27 & 0.37 & 0.22 & 0.18 & 0.41 & 0.33 & 0.38 & 0.39 & fail & fail \\
        \hline
        FOM(mean) & 3.47 & \textbf{3.43} & 3.52 & 3.49 & 3.75 & 3.65 & 3.95 & 4.00 & 11.80 & 11.85 & fail & fail \\
        FOM(median) & 3.44 & 3.44 & 3.53 & \textbf{3.40} & 3.70 & 3.55 & 3.97 & 4.99 & 11.67 & 12.31 & fail & fail \\
        FOM(best) & 3.29 & \textbf{3.22} & 3.26 & 3.28 & 3.50 & 3.28 & 3.48 & 3.74 & 8.26 & 9.29 & fail & fail \\
        FOM(worst) & 3.77 & \textbf{3.61} & 3.95 & 3.88 & 4.16 & 4.30 & 4.48 & 4.43 & 14.03 & 13.40 & fail & fail \\
        \hline
        Avg. \# Sim & 131 & 139 & 73 & 72 & 44 & 50 & 790 & 2328 & 1599 & 1538 & fail & fail \\
        \# Success & 12/12 & 12/12 & 12/12 & 12/12 & 12/12 & 12/12 & 12/12 & 12/12 & 12/12 & 12/12 & 0/12 & 0/12 \\
        \hline
    \end{tabular}
    \vspace{-0.1in}
\end{table*}

In this experiment, we randomly sample 20 initial data points and set the maximum number of simulations as 920 for both MACE and \textit{o}MACE, regardless of the batch size. For WEIBO, we initially sample 20 data points and limit the simulation budget as 720. For GASPAD, the simulation budget is limited as 1000. For the rest of the algorithms, we fix the maximum number of simulations as 3000. To ensure a fair comparison, we run each algorithm 12 times to reduce the random fluctuations. The constrained optimization results of the low-power three-stage amplifier circuit are presented in Table \ref{tb:three_amplifier}. 


In this experiment, both SA and MSP fail to meet the design specification in all runs. Compared with WEIBO, GASPAD, DE and PSO, MACE reduce the simulation time consumption by up to $6\times$, $13\times$, $42\times$ and $42\times$ respectively, while achieving higher sampling efficiency. The fact that both MACE and \textit{o}MACE consistently outperform the state-of-the-art optimization algorithms again confirms the efficiency and effectiveness of our proposed batch policy. Another interesting observation is that the performance of \textit{o}MACE deteriorates quickly with the increase of the batch size, while MACE achieves relatively the same optimization results across different batch sizes. This clearly demonstrates the robustness and effectiveness of the two-stage approach.

\begin{figure}
\centering
\includegraphics[width=0.45\textwidth]{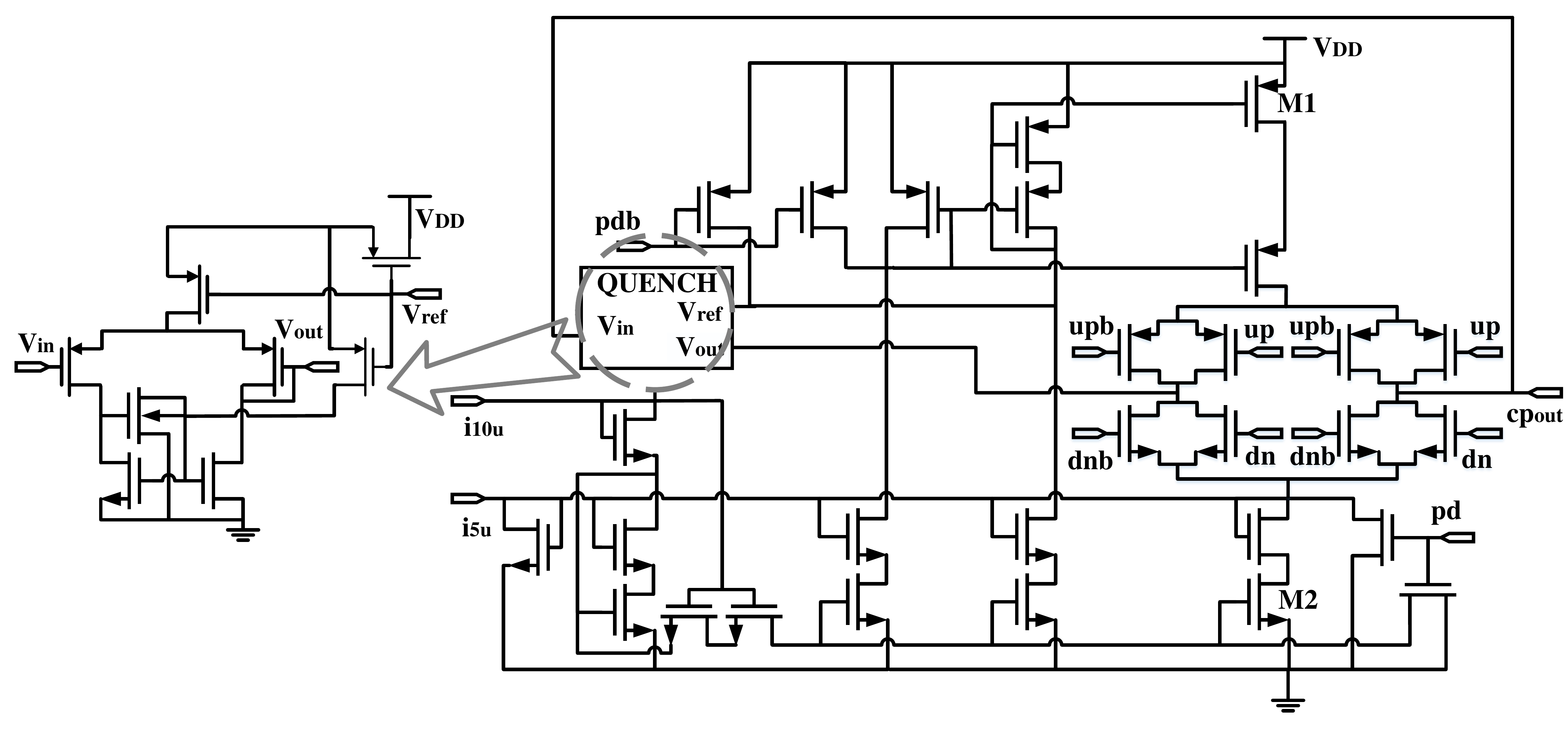}
\vspace{-0.1in}
\caption{The schematic of the charge pump circuit, which is reproduced from \cite{yang2018smart-msp:}.\vspace{-5mm}}
\label{fig:charge_pump}
\end{figure}

\subsection{Charge Pump}

The last circuit we evaluate MACE with is the charge pump circuit, the schematic of which is presented in Figure \ref{fig:charge_pump}. Implemented in a SMIC 40nm process, there are a total of 36 design variables. In this circuit, our target of design is to let the current difference between M1 and M2 stay under $40\mu$A. For this circuit, we only run experiments on the constrained optimization problem.

\textbf{Constrained optimization.} The formulated design specification for the constrained optimization problem is as follow:
\begin{equation}
    \label{eq:charge_pump1}
    \begin{aligned}
        \text{minimize} \quad & FOM & & \\
        \text{s.t.} \quad & \mathit{diff_1} & < \quad & 20\mu A, \\
        \quad & \mathit{diff_2} & < \quad & 20\mu A, \\
        \quad & \mathit{diff_3} & < \quad & 5\mu A, \\
        \quad & \mathit{diff_4} & < \quad & 5\mu A, \\
        \quad & \mathit{deviation} & < \quad & 5\mu A,
    \end{aligned}
\end{equation}
where
\begin{equation}
    \label{eq:charge_pump2}
    \begin{cases}
        \begin{aligned}
            \mathit{diff_1} \quad & = \quad \text{max}_{\forall PVT} (I_{M1, max} - I_{M1, avg}), \\
            \mathit{diff_2} \quad & = \quad \text{max}_{\forall PVT} (I_{M1, avg} - I_{M1, min}), \\
            \mathit{diff_3} \quad & = \quad \text{max}_{\forall PVT} (I_{M2, max} - I_{M2, avg}), \\
            \mathit{diff_4} \quad & = \quad \text{max}_{\forall PVT} (I_{M2, avg} - I_{M2, min}), \\
            \mathit{diff}   \quad & = \quad \sum^4_{i=1} \mathit{diff_i}, \\
            \mathit{deviation} \quad & = \quad \text{max}_{\forall PVT} (\lvert I_{M1, avg} - 40\mu A \rvert) \\
            \quad & + \quad \text{max}_{\forall PVT} (\lvert I_{M2, avg} - 40\mu A \rvert), \\
            FOM \quad & = \quad 0.3 \times \mathit{diff} + 0.5 \times \mathit{deviation}.
        \end{aligned}
    \end{cases}
\end{equation}

In this experiment, we run both MACE and \textit{o}MACE in 3 different batch sizes and compare experimental results with WEIBO, GASPAD, DE, PSO, and SA. For MACE and \textit{o}MACE, we set the size of the initial dataset as 20 and fix the simulation budget as 830, regardless of the batch size. For WEIBO, we initially sample 120 data points and limit the overall simulation budget as 1000. For the rest of the algorithms, we fix the maximum number of simulations as 2500. Again, to ensure a fair comparison, we run each algorithm 12 times to average the random fluctuations. The corresponding mean, median, best-case and worst-case optimization results are presented in Table \ref{tb:charge_pump}. 


In this experiment, PSO and SA fail to find feasible designs in all 12 runs. Compared with WEIBO, GASPAD, MSP and DE, MACE again achieves much better optimization results while reducing the simulation time by $15\times$, $46\times$, $31\times$ and $30\times$. This shows that MACE has much higher convergence rate and explore the design space more efficiently than the state-of-the-art optimization algorithms. One interesting observation is that GASPAD achieves worse optimization results compared to WEIBO while requiring much more simulation time on average. This indicates that the response surface of the corresponding design specification is multi-modal and requires a large set of data points to approximate. The fact that MACE consistently outperforms \textit{o}MACE across different batch sizes again suggests that the two-stage policy helps to provide more information about the feasible region and can greatly accelerate the optimization procedure.

\section{Conclusion} \label{sec:conclusion}

In this paper, we propose a batch Bayesian optimization algorithm based on the acquisition function ensemble. Our algorithm can handle both unconstrained and constrained optimization problems. By sampling data points from the Pareto front of PI, EI and LCB, we combine the benefits of state-of-the-art acquisition functions and achieve a delicate tradeoff between exploration and exploitation for the unconstrained optimization problem. Fueled with this explicitly designed batch policy, we further refine the algorithm to handle the constrained optimization problem by dividing the optimization procedure into two stages. By first focusing on finding the feasible designs, we manage to collect more information about the feasible region. We further reduce sampling around the invalid region while exploring the potential area by adopting a carefully designed penalization term. The experimental results demonstrate the robustness and cost-effectiveness of our proposed algorithm.


\ifCLASSOPTIONcaptionsoff
  \newpage
\fi

\bibliographystyle{IEEEtran}
\bibliography{include/bibliography}

\vspace{-0.5in}
\begin{IEEEbiography}[{\includegraphics[width=1in, clip, keepaspectratio]{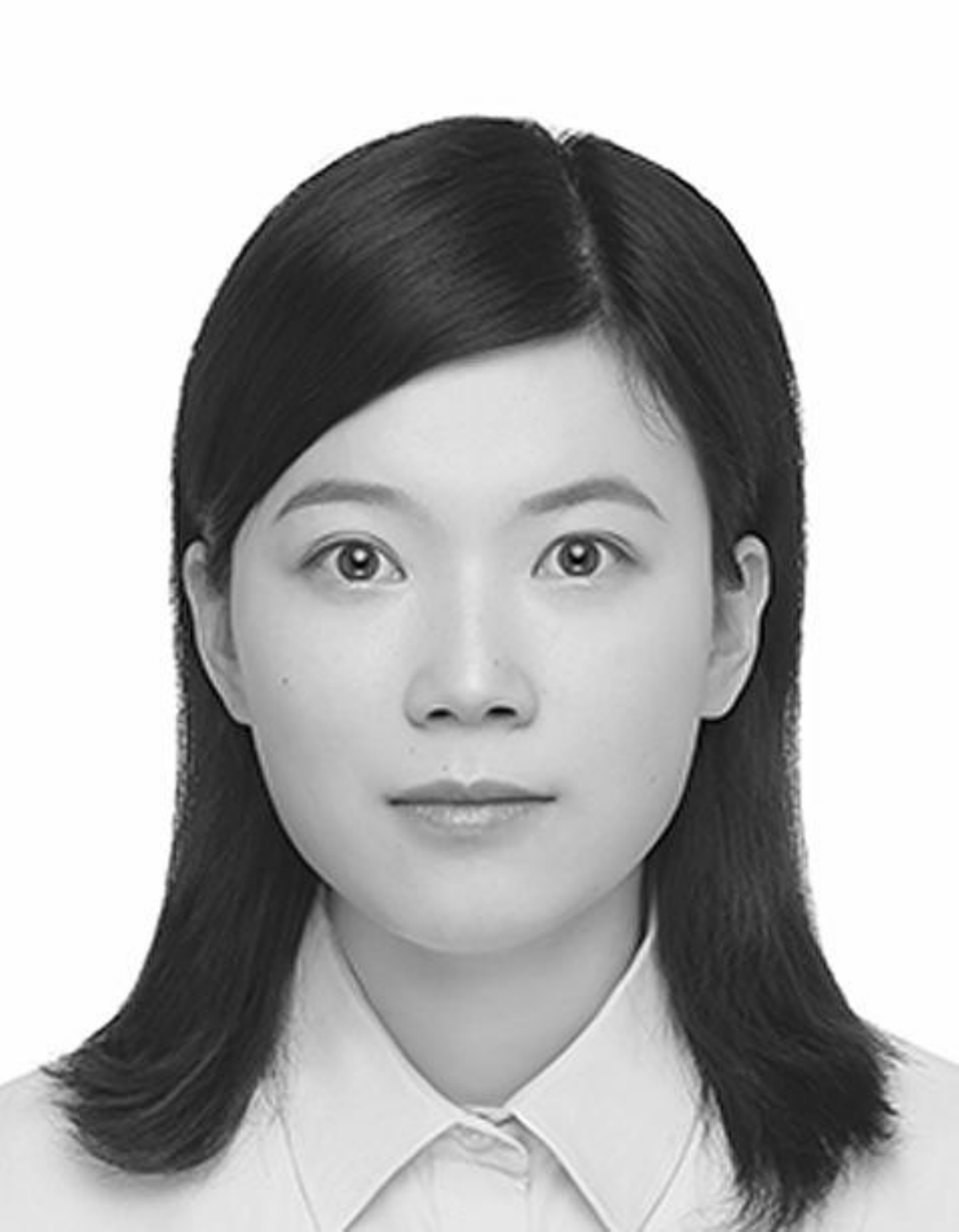}}]
    {Shuhan Zhang} received the B.S. degree in microelectronics from Fudan University, Shanghai, China in 2016. 
    
    She is currently pursuing the Ph.D. degree with State Key Lab. of Application Specific Integrated Circuits and System, Microelectronics Department, Fudan University, Shanghai, China. Her current research interests include analog circuit design automation and optimization, few-shot learning, generative adversarial networks and style transfer.
\end{IEEEbiography}

\vspace{-0.5in}
\begin{IEEEbiography}[{\includegraphics[width=0.9in, clip, keepaspectratio]{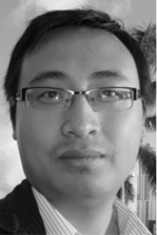}}]
{Fan Yang} (M’08) received the B.S.~degree from Xi’an Jiaotong University in 2003, and the Ph.D.~degree from Fudan University in 2008. From 2008 to 2011, he was an Assistant Professor with Fudan University. He is currently an Associate Professor with the Microelectronics Department, Fudan University. His research interests include model order reduction, circuit simulation, high-level synthesis, yield analysis, and design for manufacturability.
\end{IEEEbiography}

\vspace{-0.5in}
\begin{IEEEbiography}[{\includegraphics[width=1in, clip, keepaspectratio]{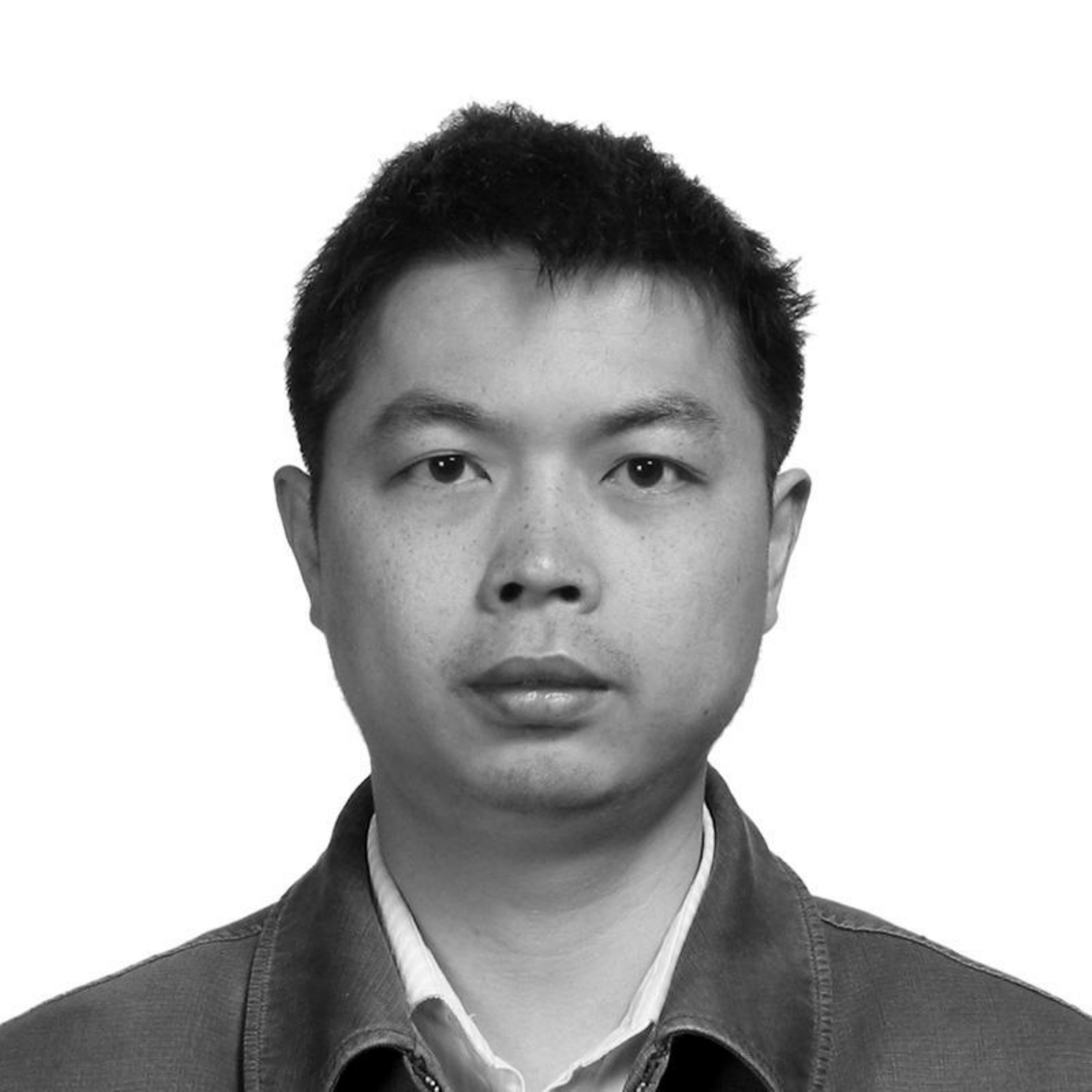}}]
    {Changhao Yan} received the Ph.D. degree in computer science from the Tsinghua University, Beijing, China, in 2006, and received the B.E. and M.E. degrees  from the Huazhong University of Science and Technology, Wuhan, China, in 1996 and 2002, respectively. He is currently an Associate Prof. of School of Microelectronics, Fudan University, Shanghai, China. His current research interests include parasitic parameter extraction of interconnects, parallel algorithms for large scale computation, design for manufacturability, robust analysis of circuits, and AI and machine learning in medicine.
\end{IEEEbiography}

\vspace{-0.5in}
\begin{IEEEbiography}[{\includegraphics[width=1in, clip, keepaspectratio]{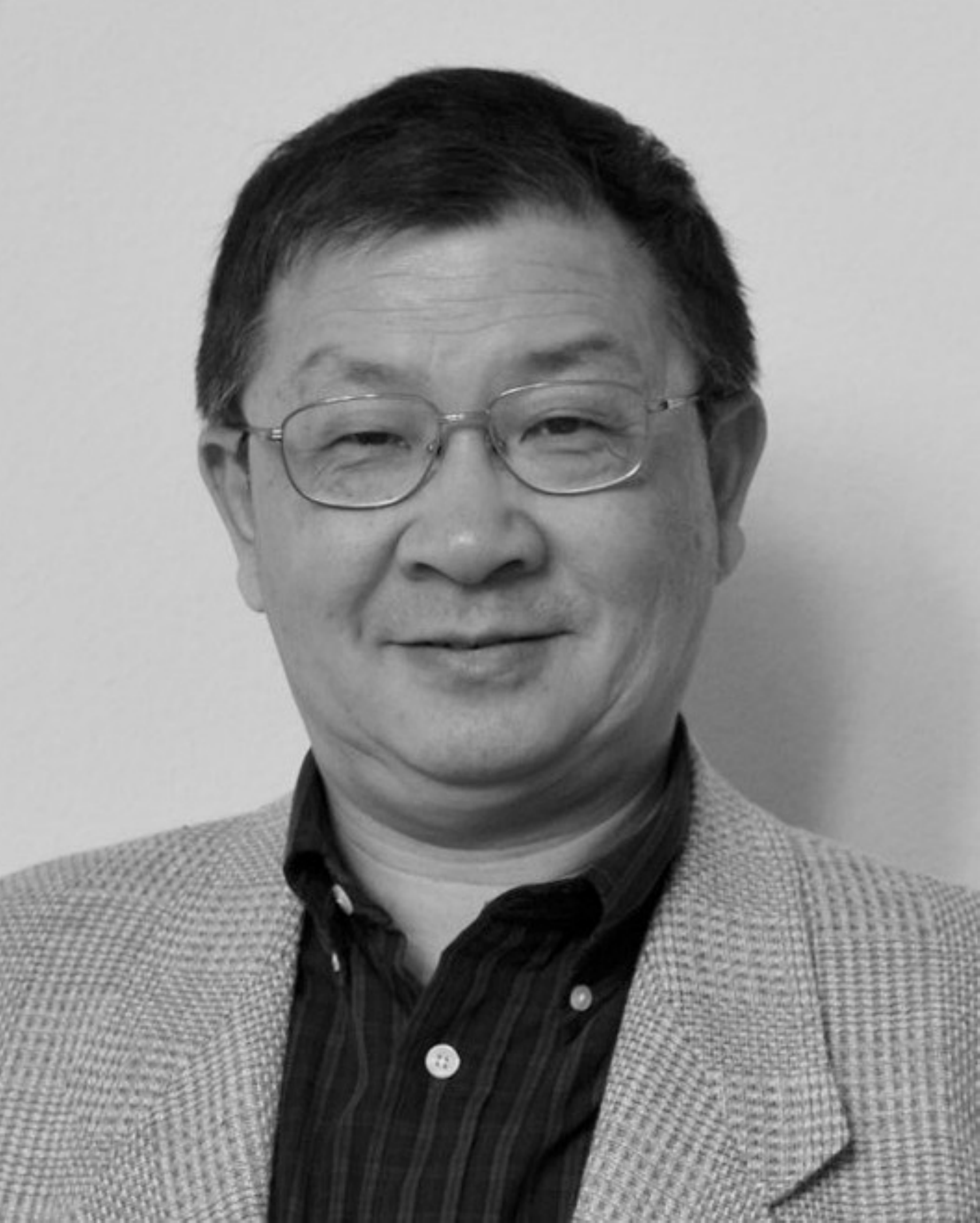}}]
    {Dian Zhou} (M’89--SM’07) received the B.S.~degree in physics and the M.S.~degree in electrical engineering from Fudan University, China, in 1982 and 1985, respectively, and the Ph.D.~degree in electrical and computer engineering from the University of Illinois in 1990. He joined the University of North Carolina at Charlotte as an Assistant Professor in 1990, where he became an Associate Professor in 1995. He joined The University of Texas at Dallas as a Full Professor in 1999. His research interests include high-speed VLSI systems, CAD tools, mixed-signal ICs, and algorithms.
\end{IEEEbiography}

\vspace{-0.5in}
\begin{IEEEbiography}[{\includegraphics[width=1in, clip, keepaspectratio]{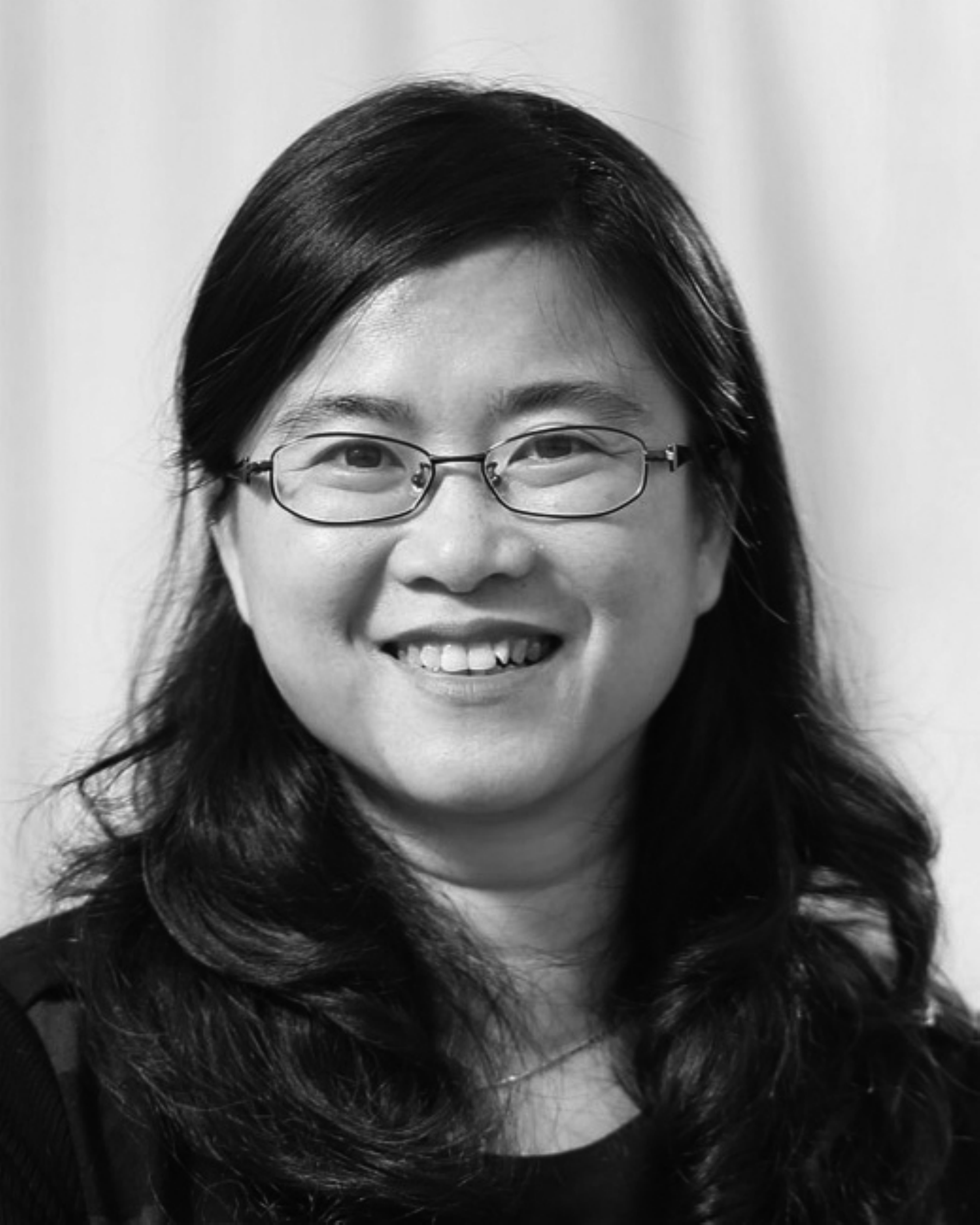}}]
    {Xuan Zeng} (M’97--SM'17)  received the B.S. and Ph.D. degrees in electrical engineering from Fudan University, Shanghai, China, in 1991 and 1997, respectively. She is a Full Professor with the Department of Microelectronics, Fudan University. She was the Director of the State Key Laboratory of ASIC \& System from 2008 to 2012. She was a Visiting Professor at the Department of Electrical Engineering, Texas A\&M University, College Station, TX, USA, and the Department of Microelectronics, Technische Universiteit Delft, Delft, The Netherlands, in 2002 and 2003, respectively. Her current research interests include analog circuit modeling and synthesis, design for manufacturability, high-speed interconnect analysis and optimization, circuit simulation. Prof. Zeng is an Associate Editor of IEEE Trans. on Circuits and Systems: Part II, IEEE Trans. on Computer-Aided Design of Integrated Circuits and Systems, and ACM Trans. on Design Automation on Electronics and Systems. Dr. Zeng received the Best Paper Award from the 8th IEEE Annual Ubiquitous Computing, Electronics \& Mobile Communication Conference 2017. She received Changjiang Distinguished Professor with the Ministry of Education Department of China in 2014, the Chinese National Science Funds for Distinguished Young Scientists in 2011 and the First-Class of Natural Science Prize of Shanghai in 2012, 10th For Women in Science Award in China in 2013, Shanghai Municipal Natural Science Peony award in 2014. 
\end{IEEEbiography}

\end{document}